\documentclass[lettersize,journal]{IEEEtran}
\usepackage{amsmath,amsfonts}
\usepackage{algorithmic}
\usepackage{algorithm}
\usepackage{array}
\usepackage[caption=false,font=normalsize,labelfont=sf,textfont=sf]{subfig}
\usepackage{textcomp}
\usepackage{stfloats}
\usepackage{url}
\usepackage{verbatim}
\usepackage{graphicx}
\usepackage{cite}

\usepackage{amssymb}
\usepackage{mathrsfs}
\usepackage{tcolorbox}
\tcbuselibrary{breakable}
\usepackage{booktabs}
\graphicspath{{./fig/}}
\usepackage{subfig}
\DeclareGraphicsExtensions{.pdf,.jpeg,.png,.eps}
\usepackage{multirow}
\usepackage{amsthm} %for align

\usepackage{booktabs}
\usepackage{xcolor}
\usepackage{mathrsfs} % 小写字母
\usepackage{amsthm,amsmath,amssymb}
\usepackage{afterpage}
\usepackage{colortbl}
\usepackage{tabularx}
\usepackage{listings}

\usepackage{hyperref}
\hypersetup{
  colorlinks=true,
  linkcolor=blue,
  citecolor=blue,
  urlcolor=blue,
}

\colorlet{punct}{red!60!black}
\definecolor{background}{HTML}{EEEEEE}
\definecolor{delim}{RGB}{20,105,176}
\colorlet{numb}{magenta!60!black}
\lstdefinelanguage{json}{
    basicstyle=\normalfont\ttfamily,
    numberstyle=\scriptsize,
    stepnumber=1,
    numbersep=8pt,
    showstringspaces=false,
    breaklines=true,
    frame=lines,
    backgroundcolor=\color{background},
    literate=
     *{0}{{{\color{numb}0}}}{1}
      {1}{{{\color{numb}1}}}{1}
      {2}{{{\color{numb}2}}}{1}
      {3}{{{\color{numb}3}}}{1}
      {4}{{{\color{numb}4}}}{1}
      {5}{{{\color{numb}5}}}{1}
      {6}{{{\color{numb}6}}}{1}
      {7}{{{\color{numb}7}}}{1}
      {8}{{{\color{numb}8}}}{1}
      {9}{{{\color{numb}9}}}{1}
      {:}{{{\color{punct}{:}}}}{1}
      {,}{{{\color{punct}{,}}}}{1}
      {\{}{{{\color{delim}{\{}}}}{1}
      {\}}{{{\color{delim}{\}}}}}{1}
      {[}{{{\color{delim}{[}}}}{1}
      {]}{{{\color{delim}{]}}}}{1},
}

\usepackage{tikz,xcolor}
\hypersetup{hidelinks,
	colorlinks=true,
	allcolors=black,
	pdfstartview=Fit,
	breaklinks=true}
	
\definecolor{lime}{HTML}{A6CE39}
\DeclareRobustCommand{\orcidicon}{
\begin{tikzpicture}
\draw[lime, fill=lime] (0,0)
circle[radius=0.13]
node[white]{{\fontfamily{qag}\selectfont \tiny \.{I}D}};
\end{tikzpicture}
\hspace{-2mm}
}
\foreach \x in {A, ..., Z}{%
\expandafter\xdef\csname orcid\x\endcsname{\noexpand\href{https://orcid.org/\csname orcidauthor\x\endcsname}{\noexpand\orcidicon}}
}
 
\begin{document}

\title{IVLMap: Instance-Aware Visual Language Grounding for Consumer Robot Navigation}

\author{Jiacui Huang, Hongtao Zhang, Mingbo Zhao, ~\IEEEmembership{Senior,~IEEE,} Wu Zhou, ~\IEEEmembership{Senior,~IEEE} 
        % <-this % stops a space
\thanks{This paper was produced by the IEEE Publication Technology Group. They are in Piscataway, NJ.}% <-this % stops a space
\thanks{*Corresponding Author}}

\markboth{IEEE TRANSACTIONS ON CONSUMER ELECTRONICS}%
{Shell \MakeLowercase{\textit{et al.}}: A Sample Article Using IEEEtran.cls for IEEE Journals}

% \IEEEpubid{Authorized licensed use limited to: Donghua University. Downloaded on January 16,2024 at 15:50:26 UTC from IEEE Xplore. Restrictions apply.}
% % column for its text to clear the IEEEpubid mark.

\maketitle

\begin{abstract}
Vision-and-Language Navigation (VLN) is a challenging task that requires a robot to navigate in photo-realistic environments with human natural language promptings. Recent studies aim to handle this task by constructing the semantic spatial map representation of the environment, and then leveraging the strong ability of reasoning in large language models for generalizing code for guiding the robot navigation. However, these methods face limitations in instance-level and attribute-level navigation tasks as they cannot distinguish different instances of the same object. To address this challenge, we propose a new method, namely, Instance-aware Visual Language Map (IVLMap), to empower the robot with instance-level and attribute-level semantic mapping, where it is autonomously constructed by fusing the RGBD video data collected from the robot agent with special-designed natural language map indexing in the bird’s-in-eye view. Such indexing is instance-level and attribute-level. In particular, when integrated with a large language model, IVLMap demonstrates the capability to \textcolor{blue}{i)} transform natural language into navigation targets with instance and attribute information, enabling precise localization, and \textcolor{blue}{ii)} accomplish zero-shot end-to-end navigation tasks based on natural language commands. Extensive navigation experiments are conducted. Simulation results illustrate that our method can achieve an average improvement of 14.4\% in navigation accuracy. Code and demo are released at https://ivlmap.github.io/.
\end{abstract}

\begin{IEEEkeywords}
Vision-and-Language Navigation, 3D Reconstruction, Zero-Shot Navigation, Visual Language Map
\end{IEEEkeywords}

\section{Introduction}

\IEEEPARstart{D}{eveloping} a robot that can cooperate with humans is of great importance in many real-world applications, such as Telsa Optimus, Mobile ALOHA\cite{fu2024mobile}, etc. A robot agent that can understand human language and navigate intelligently is therefore significant to benefit human society. Towards this end, Vision-and-Language Navigation (VLN) has been developed during the past few years, which is to empower a robot agent to navigate in photo-realistic environments according to natural language instructions, such as "navigate to the 3rd chair" or "navigate to the yellow sofa" \cite{anderson2018vision,gu2022vision}. This requires the robot agent can interpret natural language from humans, perceive its visual environment, and utilize the information to navigate, where a key important issue is how to structure the visited environment and make global planning. To handle this case, a few recent approaches \cite{shah2023lm} utilize the topological map for structure. But these methods are difficult to represent the spatial relations among objects, causing detailed information may be lost; on the other hand, more recent works \cite{huang2023visual} model the navigation environment using the top-down semantic map, which represents spatial relations more precisely. However, the semantic concepts are extremely limited due to the pre-defined semantic labels.

In general, based on the environments for navigation, the VLN can be roughly divided into two categories, which include navigation in discrete and continuous environments. In discrete environments, such as those in R2R \cite{anderson2018vision}, REVERIE \cite{qi2020reverie}, and SOON\cite{zhu2021soon}, the VLN is conceptualized as a topology structure comprised of interconnected navigable nodes. The agent utilizes a connectivity graph to move between adjacent nodes by selecting a direction from the available navigable directions; in contrast, VLN in continuous environments, such as those in R2R-CE\cite{wang2023scaling} and RxRCE\cite{he2021landmark}, enable the agents to have the flexibility to navigate to any unobstructed point using a set of low-level actions (e.g., move forward 0.25m, turn left 15 degrees) instead of teleporting between fixed nodes. This approach is closer to real-world robot navigation, posing a more intricate challenge for the agent.

\begin{figure*}[htp]    
\centerline{
\subfloat{\includegraphics[width=0.33\textwidth]{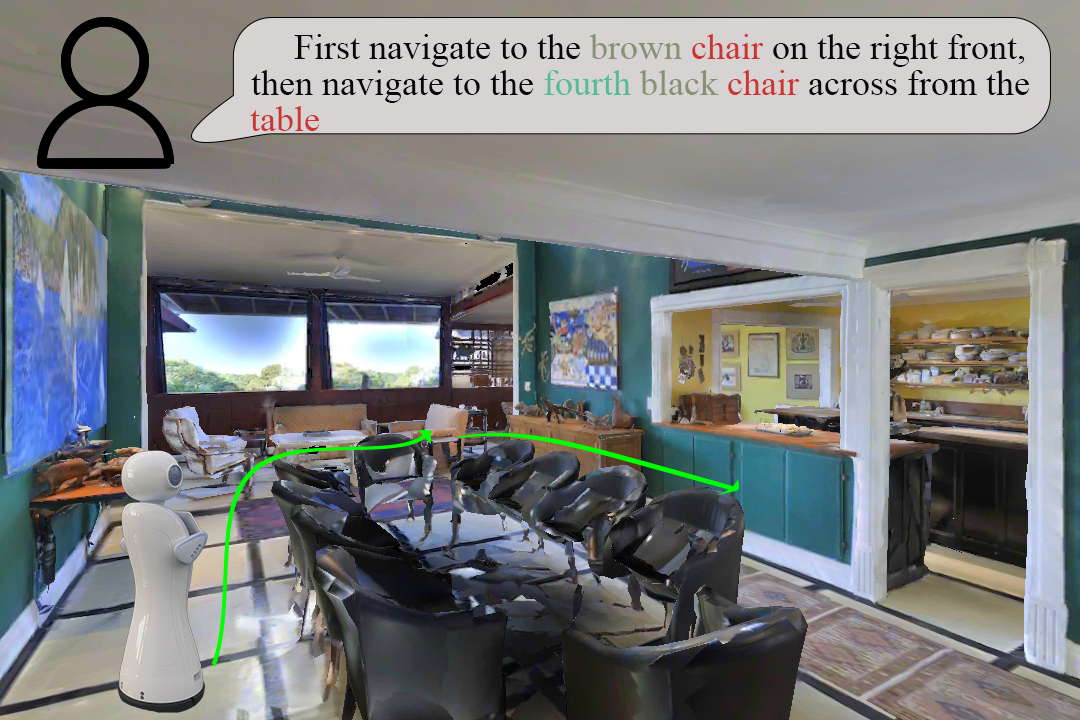}}\
\subfloat{\includegraphics[width=0.33\textwidth]{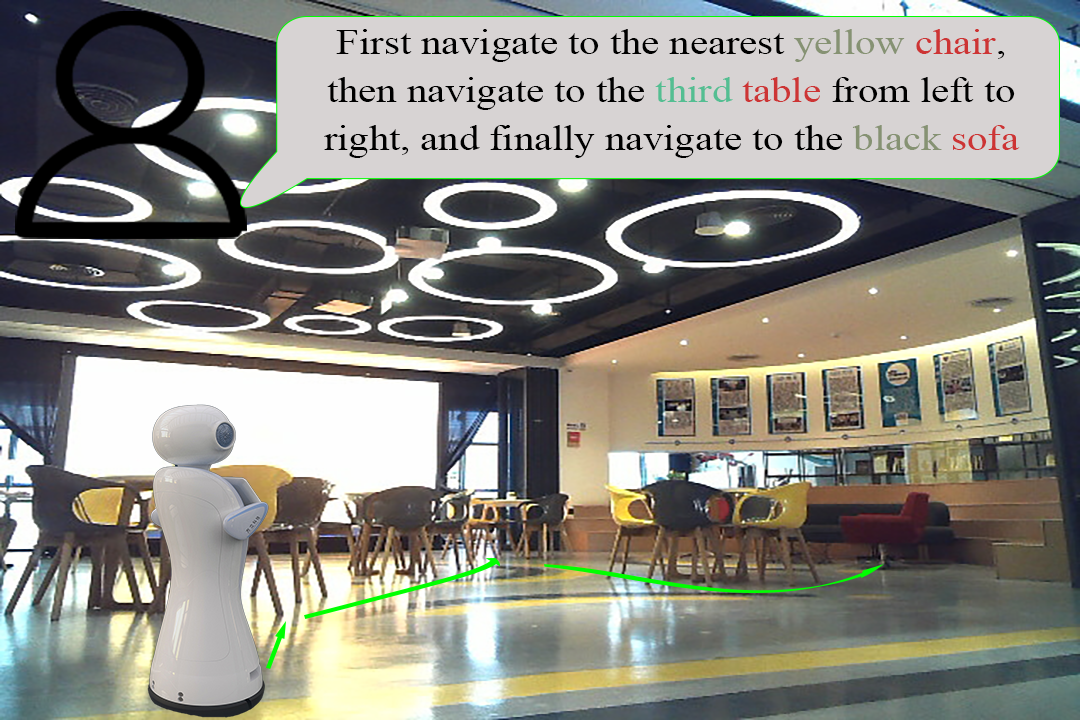}}\
\subfloat{\includegraphics[width=0.33\textwidth]{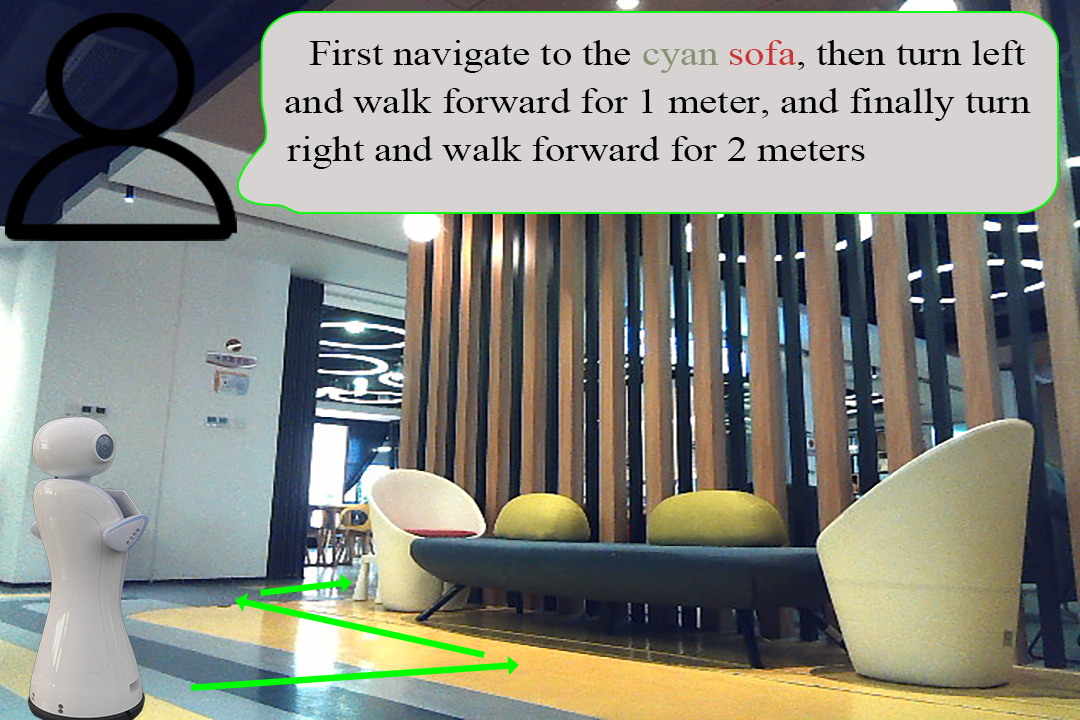}}}
\caption{Under the guidance of our IVLMap, robotic agents can accomplish instance-level target navigation. Leveraging the map's information, the robot can navigate to specific objects based on their instance attributes(color, shape, etc). This capability enables the robotic agent to execute navigation tasks with a higher degree of precision, enhancing its ability to reach designated objects accurately.}
\label{fig: navigation demonstration}
\end{figure*}

Take the instance illustrated in \textcolor{red}{Fig.\ref{fig: navigation demonstration}}, where the language instruction is to "navigate to the fourth black chair across from the table". To execute this command, we initially explore the entire room to identify all instances of tables, extracting their color attributes. Subsequently, we locate the position of the fourth table with the specified black color mentioned in the command. Completing such a navigation task becomes easily achievable if we have a global map of the scene, and the map should encompass information about each object, including its category, details, color, etc. Recently, VLMap \cite{huang2023visual} pioneers innovative zero-shot spatial navigation by constructing Semantic maps via indexing visual landmarks using natural language. However, VLMap is limited to navigating to the vicinity of the closest category to the robotic agent and cannot fulfill the envisioned, more common, and precise instance-level and attribute-level navigation needs in real-life scenarios.
%Such comprehensive data simplifies the execution of the navigation task. Researchers are developing innovative approaches, such as VLMaps, which enable robots to index visual landmarks using natural language, facilitating tasks like landmark localization\cite{huang2023visual}. VLMaps pioneers innovative zero-shot spatial navigation. With VLMaps, users achieve end-to-end natural language navigation by simply describing the object category in their surroundings using natural language. However, VLMap is limited to navigating to the vicinity of the closest category to the robotic agent and cannot fulfill the envisioned, more common, and precise instance-level and attribute-level navigation needs in real-life scenarios. 

In this paper, to address this challenge, we propose a new method, namely, Instance-aware Visual Language Map (IVLMap), to empower the robot with instance-level and attribute-level semantic mapping, where the IVLMap is autonomously constructed by fusing the RGBD video data with a specially-designed natural language map indexing in the bird’s-in-eye view. Such indexing is instance-level and attribute-level. In this way, IVLMap can well separate different instances within the same category. When integrated with a large language model,  it demonstrates the capability to \textcolor{blue}{i)} transform natural language into navigation targets with instance and attribute information, enabling precise localization, and \textcolor{blue}{ii)} accomplish zero-shot end-to-end navigation tasks based on natural language commands. The main contributions of the proposed work are as follows:
\begin{enumerate}
%\item By ingeniously acquiring the instantiation details for each object, we enable our map to achieve instance-level precise navigation. Building upon this, we introduce the Instance Level Visual Language Map (\textcolor{red}{IVLMap}).
%\item Employing a technique involving region matching and label scoring, we obtain the category label for each SAM\cite{kirillov2023segment}-segmented mask. Innovatively, using a similar approach, we acquire the color label for each mask.
\item a novel instance-level and attribute-level map in bird's-in-eye review is developed for better comprehend the environment and robot navigation planning, where the instances in map are obtained by involving region matching and label scoring to achieve the category label for each SAM\cite{kirillov2023segment}-segmented mask, while the color attribute labels for each mask are also obtained by using the similar approach.
\item leveraging IVLMap, we propose a two-step method for locating landmarks. This involves an initial coarse-grained localization of the corresponding mask, followed by a fine-grained localization and navigation within the mask's designated region.
\item we established an interactive data collection platform, achieving real-time controllable data acquisition. This not only reduced data volume but also enhanced reconstruction efficiency. Furthermore, we conducted experiments in a real-world environment, providing valuable insights for the practical deployment of this method.

\end{enumerate}

The rest of this work is organized as follows: In Section II, we will review some related work; in Section III, we will give the detailed description for presenting the proposed IVLMAP. Extensive simulations are conducted in Section IV and the final conclusions are drawn in Section V.

\section{Related Work}

\textbf{Semantic Mapping.} Advances in semantic mapping, driven by the integration of Convolutional Neural Networks (CNNs)\cite{fukushima1980neocognitron} and dense Simultaneous Localization and Mapping (SLAM)\cite{estefo2019robot}, have progressed significantly. SLAM++\cite{salas2013slam++} introduced an object-oriented SLAM approach leveraging pre-existing knowledge. Subsequent studies, e.g.,\cite{mccormac2018fusion++}, use Mask-RCNN for instance-level semantics in 3D volumetric maps.  VLMaps\cite{huang2023visual} and NLMaps-Saycan\cite{chen2023open} introduce scene representations queryable through natural language, using Visual Language Models (VLMs). While prior studies focus on semantic details, our work emphasizes pixel-level segmentation accuracy, categorizing each pixel precisely. 

% \subsection{Instance Segmentation}
% \textbf{Instance Segmentation: }The significance of achieving instance-level segmentation plays a vital role in VLN, emphasizing the crucial need to identify and precisely locate distinct instances of similar objects. Previous research\cite{zhang2019pose2seg} innovatively addresses human instance segmentation, removing the necessity for distinct detection stages. Subsequent work\cite{lee2020centermask} introduces a real-time instance segmentation method by incorporating a Spatial Attention-Guided Mask (SAG-Mask) branch. Most recently, Segment Anything Model(SAM)\cite{kirillov2023segment} developed by Meta AI, stands out due to its promptable segmentation capabilities, enabling zero-shot generalization to unfamiliar objects and images without the need for additional training, overcoming the limitations of traditional segmentation models by offering unparalleled flexibility and adaptability. However, the SAM model can only achieve object mask segmentation and is unable to provide labels for each mask, posing certain limitations for practical applications. 

\textbf{Instance Segmentation.} Achieving instance-level segmentation is crucial in VLN, requiring precise identification and location of individual instances of similar objects. Innovative research\cite{zhang2019pose2seg} addresses human instance segmentation without distinct detection stages. Real-time solutions\cite{lee2020centermask} introduce a Spatial Attention-Guided Mask (SAG-Mask) branch. Meta AI's Segment Anything Model (SAM)\cite{kirillov2023segment} excels in promptable segmentation, enabling zero-shot generalization to unfamiliar objects. SAM's object mask segmentation, while powerful, lacks individual mask labeling, limiting practical applications. 

% \subsection{Vision-and-Language Navigation(VLN)}
% \textbf{Vision-and-Language Navigation(VLN): }In recent years, the field of VLN has witnessed a thriving development, with researchers contributing significantly to the domain. Y Long and collaborators employed a collaborative approach, leveraging large-scale training to endow extensive domain knowledge and capabilities to large models\cite{long2023discuss}. The VIM-Net model proposed in this paper\cite{jia2022vital} overcomes the inability of current fusion methods to work accurately on important visual features by critically matching visual and linguistic information. Additionally, advancements like Talk2Nav focus on long-range Vision-and-Language Navigation, incorporating dual attention mechanisms and spatial memory for improved navigational conceptualization\cite{vasudevan2021talk2nav}. At the forefront of VLN research, several challenges endure. These encompass the incapability to navigate through unseen objects\cite{huang2023visual}, the precision constraints in interpreting language descriptions\cite{sun2023outdoor}, the difficulty in personalizing navigation tailored to object attributes\cite{chen2022weakly}, and the demanding data-intensive requirements for achieving end-to-end navigation\cite{kolve2017ai2}.

\textbf{Vision-and-Language Navigation(VLN).} Recent years have seen significant strides in VLN, driven by researchers such as Y. Long et al., who employed large-scale training to imbue extensive domain knowledge in models like VIM-Net\cite{long2023discuss}. Addressing fusion challenges, VIM-Net critically matches visual and linguistic information for accurate navigation\cite{jia2022vital}. Innovations like Talk2Nav, with dual attention mechanisms and spatial memory, tackle long-range VLN\cite{vasudevan2021talk2nav}. Challenges persist, including navigating unseen objects, interpreting language descriptions precisely, personalizing navigation for object attributes, and managing data-intensive requirements for end-to-end navigation\cite{huang2023visual,chen2022weakly,krantz2023iterative}. 

% \subsection{LLM used in VLN}
% \textbf{LLM used in VLN: }Enhanced statement on language models in robotics and aavigation, Large Language Models (LLMs) play a pivotal role in VLN tasks, as demonstrated by the introduction of NavGPT, a navigation agent explicitly showcasing the reasoning capabilities of GPT\cite{zhou2023navgpt}. LM-Nav further underscores the significance of LLMs by employing GPT-3 to parse instructions and navigate based on landmarks\cite{shah2023lm}. VELMA expands LLM capabilities to real-world scenarios, such as street-level navigation, utilizing verbalization techniques to articulate trajectory and environment observations. This contribution enriches the landscape of embodied artificial intelligence for VLN in complex urban environments\cite{schumann2023velma}. Vemprala et al explore a strategy to enhance ChatGPT's proficiency in solving robotics tasks, aiming to establish a framework for seamlessly integrating language models into robotics applications\cite{vemprala2023chatgpt}. Additionally, PaLM-E represents a cutting-edge advancement in embodied language models, seamlessly integrating visual and textual inputs for multimodal reasoning tasks\cite{driess2023palm}. Our project is influenced by the studies mentioned, and we have successfully implemented accurate navigation using LLMs guided by natural language. 

\textbf{LLM used in VLN.} Large Language Models (LLMs) play a pivotal role in VLN, evident in NavGPT's reasoning capabilities\cite{zhou2023navgpt}. LM-Nav uses GPT-3 for parsing instructions and navigating landmarks\cite{shah2023lm}. VELMA extends LLMs to street-level navigation, enriching AI in complex urban environments\cite{schumann2023velma}. Studies like Vemprala et al.'s integration of ChatGPT into robotics applications and PaLM-E's multimodal reasoning with visual and textual inputs\cite{vemprala2023chatgpt,driess2023palm} influence our project, where we achieved accurate navigation using LLMs guided by natural language.

\section{Method}
\begin{figure*}[ht]
    \centering
    \includegraphics[width=0.74\textwidth]{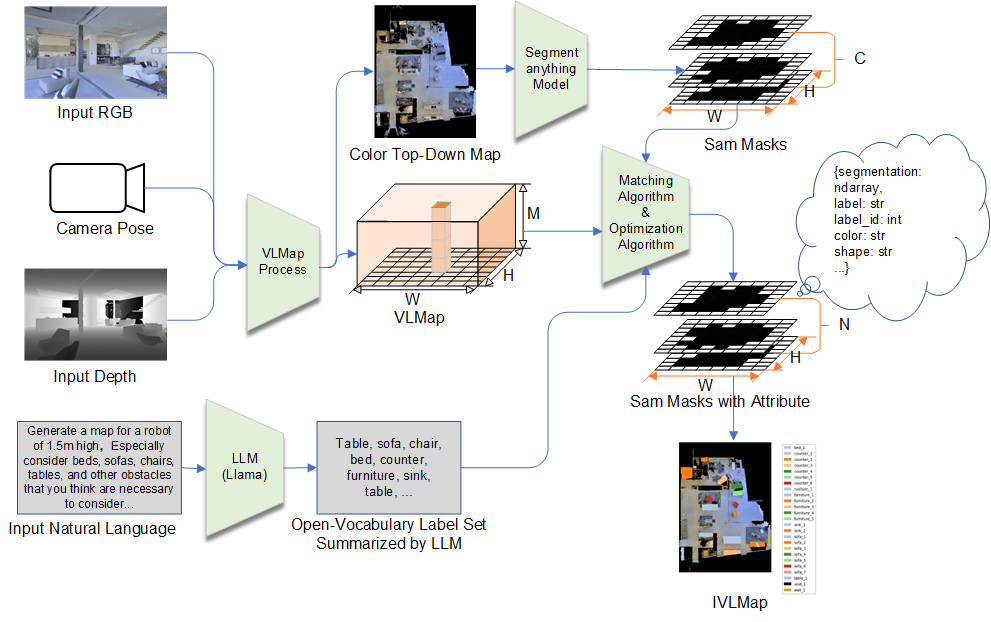}
    \caption{The IVLMap pipeline consists of two main components. The first focuses on 3D reconstruction and constructing a visual language map. Building on this foundation, the second part integrates the Segment Anything Model (SAM), enhancing the map's representation with a segmentation-aware approach for more detailed information.}
    \label{ivlmap pipeline}
\end{figure*}

In our work, our objective is to construct a semantic map of the surrounding environment which also encompasses instance-level information and attribute information of objects.  It is crucial to emphasize the inclusion of instance-level semantic information and object attributes in maps. This incorporation is vital for effectively processing linguistic commands commonly utilized in everyday language. In everyday life, robots commonly encounter commands like "navigate to the fourth black chair across from the table". Robots are required to identify "which of the chair" both in terms of its sequence and attributes. Simultaneously, they are required to find "where is the chair" by paying attention to the spatial relationship between the chair and the table.  Our method is proposed based on VLMap\cite{huang2023visual}. In the following subsections, we describe (i) how to build IVLMap (\textcolor{red}{Sec.\ref{section: ivlmap-creation}}), (ii) how to use these maps to localize open-vocabulary landmarks (\textcolor{red}{Sec.\ref{section: open-vocalbulary}}), (iii) How to employ VLMaps in tandem with large language models (LLMs) to enable instance level  object goal navigation using natural language commands (\textcolor{red}{sec.\ref{section: zero-shot-navegation}}). Our pipeline is visualized in \textcolor{red}{Fig.\ref{ivlmap pipeline}}. 

\subsection{IVLMap Creation}
\label{section: ivlmap-creation}
\textbf{Construction of VLMap and 3D Reconstruction Map in Bird's-Eye View:} The fundamental concept of VLMap involves integrating pre-trained visual language features into a three-dimensional (3D) reconstruction map.  In our work, we utilize LSeg\cite{Li_Weinberger_Belongie_Koltun_Ranftl} as the visual-language model,  a language-driven semantic segmentation model that segments the RGB images based on a set of free-form language categories. As described in VLMap, VLMap is defiened as  ${\cal M} \in {{\mathbb R}^{\bar H \times \bar W \times C}}$, where ${\bar H}$ and ${\bar W}$ represent the size of the top-down grid map, and $C$ represents the length of the VLM embeddings vector for each grid cell.  In the matrix ${\cal M}$, each cell encapsulates category information corresponding to pixels, which will furnish semantic support for the construction of our IVLMap. 

In order to get  the 3D Reconstruction Map in Bird's-Eye View, we,  in line with the approach employed by VLMaps, obtain the corresponding coordinates of each pixel u in the RGB-D data frame within the grid map by formula
\begin{equation}
    p_{{\rm{map }}}^x = \left\lfloor {\frac{{\bar H}}{2} + \frac{{P_W^x}}{s} + 0.5} \right\rfloor ,p_{{\rm{map }}}^y = \left\lfloor {\frac{{\bar W}}{2} - \frac{{P_W^z}}{s} + 0.5} \right\rfloor
    \label{eq:1}
\end{equation}
where ${{\bf{P}}_k} \in {{\mathbb R}^3}$ is the pixel's 3D point position in the k-th frame, ${{\bf{P}}_w} \in {{\mathbb R}^3}$ is the pixel's 3D point position in the world coordinate frame, $p_{{\rm{map }}}^x$ and $p_{{\rm{map }}}^y$ represent the coordinates of the projected point in the map ${\cal M}$.  We define the 3D Reconstruction Map in Bird's-Eye View as ${\cal B} \in {{\mathbb R}^{\bar H \times \bar W \times 3}}$, where each cell $\beta _{ij}$ at the coordinates of  $\left ( p_{{\rm{map }}}^x,  p_{{\rm{map }}}^y\right ) $ denotes the color of the corresponding position in RGB image $\mathcal{I}_{k}$. In the context of navigation, robots only need to focus on obstacles that are approximately at their height or lower (assuming the height of the robot is denoted as $h$ ). Objects significantly taller than the robot, such as ceilings or ceiling fans, do not require much attention. Intuitively, there exist multiple 3D points projecting to the same grid location in the map. We define matrix ${\cal H} \in {{\mathbb R}^{\bar H \times \bar W \times 1}}$  to represent the projection height of each pixel in every cell, it continuously updates with the processing of RGB-D data. Only pixels with a height lower than the robot's height $h$ are considered. The descending update method is employed, causing the values in each cell of ${\cal H}$ to gradually decrease during the reconstruction process. This ensures that all objects observed by the robot are taken into account in 3D Reconstruction Map ${\cal B}$.

\textbf{Integrating Instance-level information to IVLMap: }Formally, IVLMap is defined as ${\mathscr M} = \left\{\mathcal{M}, \mathcal{U}, \mathcal{V}\right\}$, where $\mathcal{M}$ represents the VLMap (\textcolor{red}{Sec.\ref{section: ivlmap-creation} Para-1}), $\mathcal{U} \in {{\mathbb R}^{\bar H \times \bar W}}$ holds instance information (represented by IDs), and $\mathcal{V} \in {{\mathbb R}^{\bar H \times \bar W}}$ contains color information. Following Kirillov et al.\cite{kirillov2023segment}, we utilize the pre-trained SAM model to segment the 3D reconstructed map ${\cal B}$, resulting in segmentation masks ${\cal \bar{S}}$, each comprising information like mask segmentation, area, and prediction accuracy. The key to achieving instance-level visual language maps is appending relevant attributes to each mask, effectively representing individual instance objects. The list of masks is denoted as $\mathcal{S} = [\mathcal{S}_{0}, \mathcal{S}_{1}, \cdots, \mathcal{S}_{N}] \in {{\mathbb R}^{\bar H \times \bar W \times N}}$, where each $\mathcal{S}{i} \in {{0,1}^{\bar H \times \bar W}}$ is the segmentation mask for the i-th instance. In the Pixel-Text Similarity Matrix $\mathrm{P}$ (\textcolor{red}{Sec.\ref{section: open-vocalbulary}}), each cell corresponds to a vector $\mathrm{P}{ij} = [p_{0}, p_{1}, \cdots, p_{N}]$ of length N, with $p_{k}$ representing the probability that the pixel at the current position belongs to the k-th category. The \textcolor{blue}{argmax\footnote{The argmax operation in PyTorch returns the index of the maximum value along a specified axis in a tensor.}} operation is applied to $\mathrm{P}$, resulting in the Pixel-Label Map $\mathrm{\bar{P}} \in {{\mathbb R}^{\bar H \times \bar W \times 1}}$, where each cell represents the label index of the corresponding pixel.

\begin{figure}[ht]
    \centering
    \includegraphics[width=0.45\textwidth]{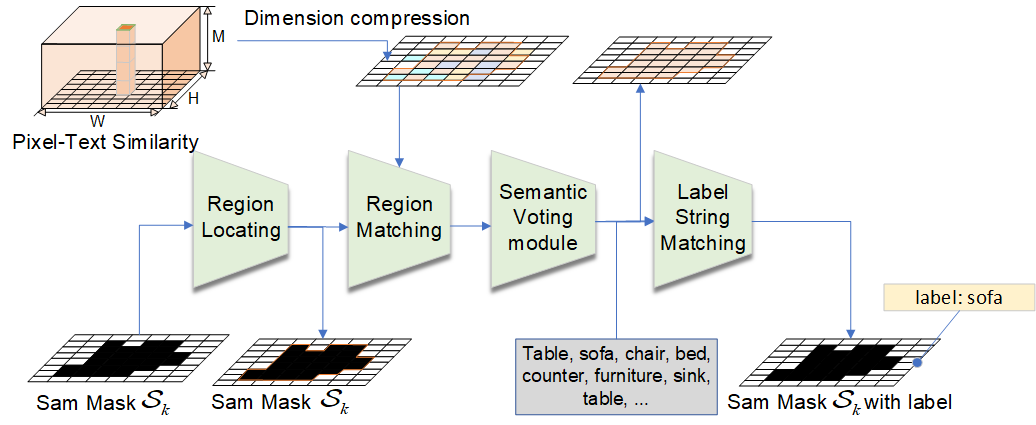}
    \caption{A schematic diagram illustrating a Matching Algorithm. We matched the masks generated by the SAM model with the Pixel-Text Similarity obtained from VLMap, assigning labels to each mask. This facilitated the subsequent implementation of IVLMap.}
    \label{fig:mask-with-label}
    \end{figure}

Due to data collection errors and model prediction inaccuracies, noisy points may exist in $\mathrm{\bar{P}}$, and instances belonging to the same category cannot be distinguished. Therefore, we employ SAM for segmenting the 3D Reconstruction Map in Bird's-Eye View and apply smoothing corrections for accurate instance segmentation. The regions in mask $\mathcal{S}{i}$ with a value of 1 represent segmented instances, and their category information is reflected in the corresponding regions of $\mathrm{\bar{P}}$. Through region matching, only the regions in $\mathrm{\bar{P}}$ that align with $\mathcal{S}{i}$ are retained, while the rest are set to zero, yielding a modified matrix $\mathrm{\bar{P'}}$. Following the approach by Chen et al, who used a top-k scoring model for obtaining category labels for SAM segmentation masks, we conduct a similar Top-k scoring operation on the category IDs in the corresponding regions of $\mathrm{\bar{P'}}$. The combination of scores determines the category of the current instance. The \textcolor{blue}{unique}\footnote{The numpy unique method efficiently identifies and returns the unique elements of an array along with the count of occurrences for each unique element, preserving their original order.} operation on $\mathrm{\bar{P'}}$ efficiently extracts unique labels $ \mathfrak{L} = [ \mathfrak{l}_{1}, \mathfrak{l}_{2}, \cdots , \mathfrak{l}_{n} ] $ and their corresponding counts $\Phi = [ \mathcal{\varphi}_{1}, \mathfrak{\varphi}_{2}, \cdots , \mathfrak{\varphi}_{n} ] $ within the matched regions, where n represents the total number of unique labels. Our scoring mechanism calculates, for each unique label within the masked region, the ratio of its count to the total number of pixels in that region, resulting in a score ranging from 0 to 1, as indicated by Equation \ref{eq:2}.
\begin{equation}
    \left\{\begin{array}{c}
S^{c}=\frac{\left[\mathfrak{\varphi}_{1}, \mathfrak{\varphi}_{1}, \cdots, \mathfrak{\varphi}_{n}\right]}{\sum \mathfrak{C}} \\
\forall i, j\left(i<j \Longrightarrow  \mathfrak{\varphi}_{i}>\mathfrak{\varphi}_{j}\right)
\end{array}\right.  
    \label{eq:2}
\end{equation}
where $S^{c} \in \mathbb{R}^n $ represents the score of each unique label. The second part of the formula signifies the descending sorting of $S^{c}$.  Additionally, the order of elements in $\mathfrak{L}$ undergoes a corresponding transformation along with the sorting of $S^{c}$. Finally, we perform the matching between category IDs and category strings. Since there is a one-to-one correspondence between $\mathfrak{L}$ and $\mathscr{C}$ (\textcolor{red}{Sec.\ref{section: open-vocalbulary}}),  we only need to match each element in $\mathfrak{L}$ to the corresponding string in  $\mathscr{C}$. The entire implementation approach is outlined in \textcolor{red}{Fig\ref{fig:mask-with-label}}. 

\textbf{Integrating additional attributes into the IVLMap: }  The essence of constructing the VLMap is to establish a correspondence between each pixel and a category label. Similarly, we can use the same approach to associate each pixel with a color label. Similar to Section \textcolor{red}{\ref{section: open-vocalbulary}}, we define the color label as $\mathscr{\bar{C}} = \left [ {\bar{c} } _{1}, {\bar{c}} _{2}, ...,{\bar{c}} _{N}  \right ]$, continuing with the same operation, we can obtain Pixel-Color-Text Similarity. Naturally, the following steps can be referred to the process outlined in \textcolor{red}{Fig.\ref{fig:mask-with-label}} to ultimately obtain the color label attributes for each mask.  As each mask corresponds to a unique object instance, the label\_id for each mask can be obtained through a unique counting id.The other attributes of the mask are relatively simple and won't be elaborated here. The final situation of each attribute is presented in \textcolor{red}{Listing \ref{mask-attributes}}. Combining the steps mentioned above, we obtain the final IVLMap ${\mathscr M}$. It integrates semantic information from the VLMap, instance information from SAM segmentation masks, and color information of instances. This comprehensive map provides convenient assistance for our navigation tasks.
\begin{lstlisting}[language=json,firstnumber=1, caption={An illustrative instance of IVLMaps' masks featuring diverse attributes in JSON format.},captionpos=b, label={mask-attributes}]
{ "segmentation": 
array([[False, ..., False], 
    [False, ..., False],  
    ..., 
    [False, ..., False], 
    [False, ..., False]]), 
"area": 1901, 
"bbox": [172, 345, 70, 28], 
"predicted_iou": 0.9936274290084839, 
"point_coords": [[232.546875, 351.03125]], 
"stability_score": 0.9828214645385742, 
"crop_box": [0, 0, 363, 478], 
"label": "floor", 
"label_id": 1, 
"num_of_same_class": 5,
"color": "yellow" } 
\end{lstlisting}

\subsection{Localizing Open-Vocabulary Landmarks}
\label{section: open-vocalbulary}
Despite VLMap's excellent open-vocabulary handling, we've further optimized it, leveraging LSeg's superior natural language-driven semantic segmentation. Precision is crucial in describing object categories during open-vocabulary implementation as LSeg's\cite{Li_Weinberger_Belongie_Koltun_Ranftl} Text encoder demands encoding potential labels. In daily life, non-experts may struggle to accurately list all object categories, emphasizing the necessity of transitioning from natural language to a precise category list. ChatGPT has established LLM's proven capability in comprehending and responding to human natural language standards\cite{ray2023chatgpt}. To achieve our objective, we employ Llama2, a cutting-edge open-source language model developed by Meta\cite{touvron2023llama}. In our work, we simply input natural language phrases(such as "Generate a map for a 1.5m high robot, especially considering items such as beds, sofas, chairs, tables, and other obstacles deemed necessary to consider") into the model. Llama, in conjunction with our contextual information (including object category labels and their sizes in the scene), combines this with the specified output specifications outlined by the prompt engineering to generate a list of object categories as output. Formally, we can define the  object categories as $\mathscr{C} = \left [ {c} _{1}, {c} _{2}, ...,{c} _{N}  \right ]$, where ${c} _{1}$ represents the i-th category in text form, and $N$ represents the total number of categories extracted by LLM. Building upon the foundation laid by Huang et al.\cite{huang2023visual}, we leverage CLIP to encode the list of category characters, resulting in an embedding matrix $\mathrm{E} \in \mathbb{R}^{M \times C}$. The embedding matrix $\mathcal{M}$ of VLMap is flattened into matrix $\mathrm{Q} \in \mathbb{R}^{\bar{M} \bar{H}  \times C}$, and the pixel-to-category similarity matrix $\mathrm{P}$ is obtained through the multiplication $\mathrm{P} = \mathrm{Q}  \ast \mathrm{E^{T} } $. Each element $\mathrm {P} _{ij} $ in the matrix represents the similarity score between a pixel and a text category, reflecting the likelihood of the pixel belonging to that specific class. The obtained object categories $\mathscr{C}$ and the category similarity matrix $\mathrm{P  }$ from the aforementioned steps are employed in the construction of IVLMap in section \textcolor{red}{\ref{section: ivlmap-creation}}. When locating a specific instance object of landmarks, we first identify the specific mask $\mathcal{S}_{i}$ from the mask set $\mathcal{S}$. Then, we locate the region information of the mask in $\mathcal{U}$ and $\mathcal{V}$ of the IVLMap $\mathscr {M}$. Finally, we pinpoint the exact instance position in the VLMap $\mathcal{M}$.

\subsection{Zero-Shot Instance Level  Object Goal Navigation from Natural Language}
\label{section: zero-shot-navegation}
In this chapter, we discuss the implementation of instance-level object goal navigation based on natural language, given a set of landmark descriptions specified by natural language instructions such as
\begin{tcolorbox}[colback=green!3!white,colframe=black!30!white]
First navigate to the nearest \textcolor{brown}{yellow} \textcolor{red}{chair}, then navigate to the \textcolor{blue}{third} \textcolor{red}{table} from left to right, and finally navigate to the \textcolor{brown}{black}  \textcolor{red}{sofa}.
\label{language-instruction}
\end{tcolorbox} 
There are numerous outstanding works in the field of zero-shot navigation, such as CoW\cite{gadre2022clip}, LM-Nav\cite{shah2023lm}, VLMap\cite{huang2023visual}. Unlike their approaches, IVLMap enables navigation to instances of objects with specific attributes, such as "the third chair on the left" or "the nearest black sofa". Particularly, we employ a large language model (LLM) to interpret the given natural language commands and decompose them into subgoals with attributes including object name, object instance information, object color\cite{ahn2022can,zeng2022socratic,shah2023lm,huang2023visual}. We have established a comprehensive set of advanced function libraries tailored for IVLMap. Building upon the understanding of natural language commands by the LLM, these libraries are invoked to generate executable Python robot code\cite{liang2023code,vemprala2023chatgpt}. Additional mathematical computations may be performed if required. Below are snippets showcasing some commands and the Python code generated by the model(input command in blue, output Python code in black), the full prompt can be referred to in  \textcolor{red}{Appendix \ref{appendix: full-prompts}}. 

Additionally, numerous large language models have emerged recently. Apart from utilizing the ChatGPT API for navigation tasks, we also achieved similar results using open-source models. We employ Llama2\cite{touvron2023llama}, an advanced open-source artificial intelligence model, to accomplish this task. We utilized the \textcolor{blue}{Llama-2-13b-chat-hf\footnote{https://huggingface.co/meta-llama/Llama-2-13b-chat-hf}} model from the Hugging Face repository. This model is fine-tuned for dialogue use cases and optimized for the Hugging Face Transformers format. To accelerate the model's inference speed, we employed GPTQ\cite{frantar2022gptq},  a one-shot weight quantization method based on approximate second-order information, to quantize the model from 8 bits to 4 bits. The experiments demonstrate an average improvement of around 30\% in the inference speed after quantization.

% 这个可以考虑和下面的比较一下
\begin{tcolorbox}[colback=green!3!white,colframe=black!30!white]
\hspace*{-1.5em} \begin{tabular}{m{22.8em}} 
\textcolor{brown}{\# navigate to the third yellow table}
\end{tabular}\\
\hspace*{-1.5em} \begin{tabular}{m{22.8em}} 
\rowcolor{blue!10} obj\_attr = agent.get\_obj\_attributes('table',3,'yellow')\\
\rowcolor{blue!10} agent.face(obj\_attr)
\end{tabular}\\
\hspace*{-1.5em} \begin{tabular}{m{22.8em}} 
\textcolor{brown}{\# Move to the red sofa then stop at the right side of it}
\end{tabular}\\
\hspace*{-1.5em} \begin{tabular}{m{22.8em}}
\rowcolor{blue!10} obj\_attr = agent.get\_obj\_attributes('sofa',0,'red') \\
\rowcolor{blue!10} agent.move\_to\_right(obj\_attr) 
\end{tabular}\\
\hspace*{-1.5em} \begin{tabular}{m{22.8em}} 
\textcolor{brown}{\# Go straight for 1 meter and then walk clockwise around the red sofa for 2 turns.}
\end{tabular}\\
\hspace*{-1.5em} \begin{tabular}{m{22.8em}}
\rowcolor{blue!10} agent.move\_forward(1)  \\
\rowcolor{blue!10}obj\_attr = agent.get\_obj\_attributes( 'sofa', 0,'red' ) \\
\rowcolor{blue!10}obj\_contour = agent.get\_specifed\_obj\_pos(obj\_attr)\\
\rowcolor{blue!10} agent.face(obj\_attr), agent.turn(-90)\\
\rowcolor{blue!10} for i in range(8):\\
\rowcolor{blue!10} \quad agent.move\_forward(obj\_contour([i\%4])\\
\rowcolor{blue!10} \quad   agent.turn(90)
\end{tabular}
 \label{zero-shot-nav}
\end{tcolorbox}

% \begin{tcolorbox}[colback=green!3!white,colframe=black!30!white]
% \textcolor{blue}{\# navigate to the third yellow table}
% \begin{lstlisting}[language=Python]
% obj_attr = agent.get_obj_attributes('table', 3, 'yellow')
% agent.face(obj_attr)

% \# Move to the red sofa then stop at the right side of it
% obj_attr = agent.get_obj_attributes('sofa', 0, 'red')
% agent.move_to_right(obj_attr)

% \# Go straight for 1 meter and then walk clockwise around the red sofa for 2 turns.
% agent.move_forward(1)
% obj_attr = agent.get_obj_attributes('sofa', 0, 'red')
% obj_contour = agent.get_specified_obj_pos(obj_attr)
% agent.face(obj_attr), agent.turn(-90)

% for i in range(8):
%     agent.move_forward(obj_contour[i % 4])
%     agent.turn(90)
% \end{lstlisting}
% \label{zero-shot-nav}
% \end{tcolorbox}

\section{Experiment}

\subsection{Experimental Setup}
We employ the Habitat simulator\cite{savva2019habitat}  alongside the Matterport3D dataset\cite{anderson2018vision} to assess performance in multi-object and spatial goal navigation tasks. Matterport3D is a comprehensive RGB-D dataset, featuring 10,800 panoramic views from 194,400 RGB-D images across 90 building-scale scenes, designed for advancing research in scene understanding indoor environments. For map creation in Habitat, we capture 13,506 RGB-D frames spanning six distinct scenes and document the camera pose for each frame, utilizing the cmu-exploration environment we established(\textcolor{red}{Sec.\ref{section: cmu-exploration}}). Due to computational constraints, our navigation experiments and the execution of the Large Language Model(Llama2) are conducted on separate servers. Llama2 operates on a server equipped with two NVIDIA RTX 3090 GPUs. As for IVLMap experiment, our experimental setup comprises an NVIDIA GeForce RTX 2080 Ti GPU with 12GB VRAM. Communication between these two servers is established using the \textcolor{blue}{Socket.IO\footnote{Socket.IO is a real-time communication protocol built on WebSocket, providing event-driven bidirectional communication for seamless integration of interactive features in web applications, official website https://socket.io/.} } protocol. 

% The dataset's particulars are delineated \textcolor{red}{Table.??????\ref{table: virtual-dataset-sceen}}. 

% \begin{table}\scriptsize
% \newcommand{\tabincell}[2]{\begin{tabular}{@{}#1@{}}#2\end{tabular}\begin{tabular}{@{}#1@{}}#2\end{tabular}\begin{tabular}{@{}#1@{}}#2\end{tabular}\begin{tabular}{@{}#1@{}}#2\end{tabular}\begin{tabular}{@{}#1@{}}#2\end{tabular}\begin{tabular}{@{}#1@{}}#2\end{tabular}\begin{tabular}{@{}#1@{}}#2\end{tabular}\begin{tabular}{@{}#1@{}}#2\end{tabular}}
% \begin{tabular}{c|c|c|c|c|c|c}
% \toprule
% \tabincell{c}{scene\\name}&\tabincell{c}{5LpN3g\\DmAk7} & \tabincell{c}{5q7pvU\\zZiYa} & \tabincell{c}{5ZKStn\\Wn8Zo} & \tabincell{c}{29hnd4\\uzFmX} & \tabincell{c}{759xd9\\YjKW5} & \tabincell{c}{ac26ZM\\wG7aT} \\ 
% \midrule
% number & 1398        & 830         & 2434        & 1363        & 2619        & 4862        \\ 
% \bottomrule
% \end{tabular}
% \caption{The Infomation of Deferent Sceen's RGB-D Datasets We Collected}
% \label{table: virtual-dataset-sceen}
% \end{table}

\textbf{Baseline:} We assess IVLMap in comparison to three baseline methods, all employing visual-language models and demonstrating proficiency in zero-shot language-based navigation:
\begin{enumerate}
\item \textbf{VLMap\cite{huang2023visual}} seamlessly integrates language and visuals, autonomously constructing maps, and excels in indexing landmarks from human instructions, enhancing language-driven robots for intuitive communication in diverse navigation scenarios with open-vocabulary mapping and natural language indexing.

\item \textbf{Clip on Wheels (CoW)\cite{gadre2022clip}} achieves language-driven object navigation by creating a target-specific saliency map using CLIP and GradCAM\cite{selvaraju2017grad}. It involves applying a threshold to saliency values, extracting a segmentation mask, and planning the navigation path based on this information.
\item \textbf{The CLIP-features-based map (CLIP Map)} serves as an ablative baseline, projecting CLIP visual features onto the environment's feature map and generating object category masks through thresholding the feature similarity.

\end{enumerate}
\textbf{Evaluation Metrics:}
Similar to previous methods\cite{schumann2022analyzing,huang2023visual} in VLN literature,  we use the standard Success Ratemetric(SR)  to measure the success ratio for the navigation task.  
We assessed our IVLMap's effectiveness by (i) presenting multiple navigation targets and (ii) using natural language commands. Success is defined as the agent stopping within a predefined distance threshold from the ground truth object.

\subsection{Dataset acquisition and 3D Reconstruction}
\label{section: cmu-exploration}
To construct a map in visual and language navigation tasks, it is crucial to acquire RGB images, depth information, and pose data from the robot or its agent while it is in motion. Common datasets on the internet, such as Matterport3D\cite{anderson2018vision}, Scannet\cite{dai2017scannet}, KITTI \cite{geiger2013vision}, may not be directly applicable to our scenario. Therefore, we undertook the task of collecting a dataset tailored to our specific requirements. Our data collection efforts were conducted in both virtual and real(\textcolor{red}{Appendix.\ref{appendix: experiment-on-real-robot}}) environments to ensure comprehensive coverage.

\begin{figure}[ht]
    \centering
    \includegraphics[width=0.45\textwidth]{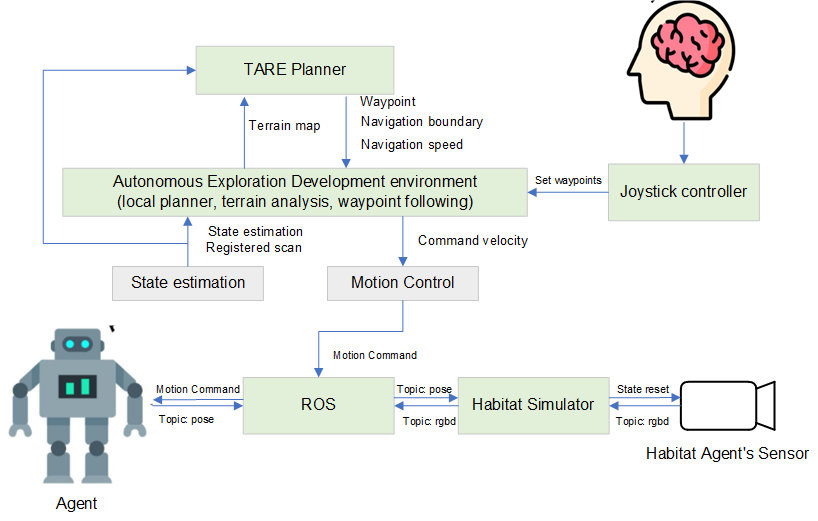}
    \caption{We created an Interactive Dataset Collection Scheme by combining the cmu-exploration development environment with the Habitat simulator. This involves integrating cmu-exploration's autonomous exploration with Habitat robot agents for a unified dataset collection approach.}
    \label{fig:dataset-base-on-vitual-environment}
    \end{figure}

\textbf{Interactive Data Collection in Virtual Habitats and CMU-Exploration Environment.} CMU-Exploration\cite{cao2022autonomous}, designed for autonomous navigation system development, offers various simulation environments and modules. Combined with the Habitat Simulator\cite{savva2019habitat}, it forms a platform where users develop and deploy navigation systems for real robots. Our interactive data collection system in this virtual environment utilizes CMU-Exploration's Joystick and Visualization Tools for waypoint setting. Waypoints undergo local planning, terrain and radar analysis, and state estimation, generating control commands for ROS-based robot motion. Simultaneously, robot pose data is sent to the Habitat simulator, providing RGB, depth, and pose information for direct sensor control. See \textcolor{red}{Fig. \ref{fig:dataset-base-on-real-environment}} for an overview. 

\begin{figure}[ht]
\centerline{\subfloat[IVLMap (Ours)]{\includegraphics[width=0.18\textwidth]{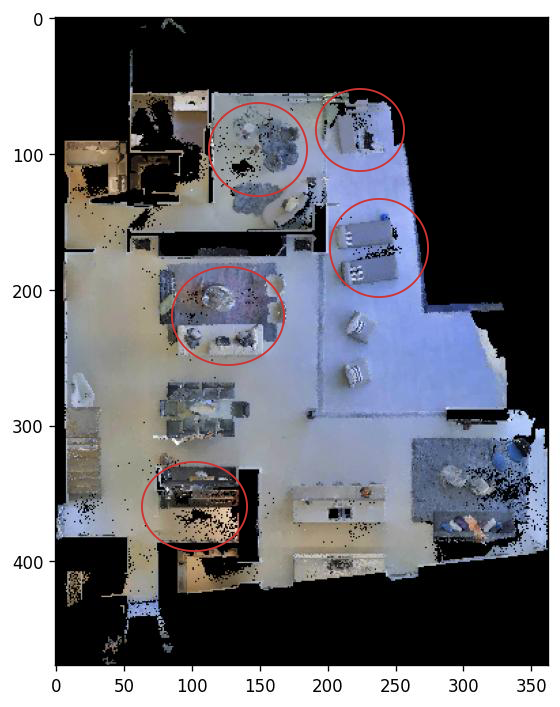}}\ \subfloat[VLMap (baseline)]{\includegraphics[width=0.18\textwidth]{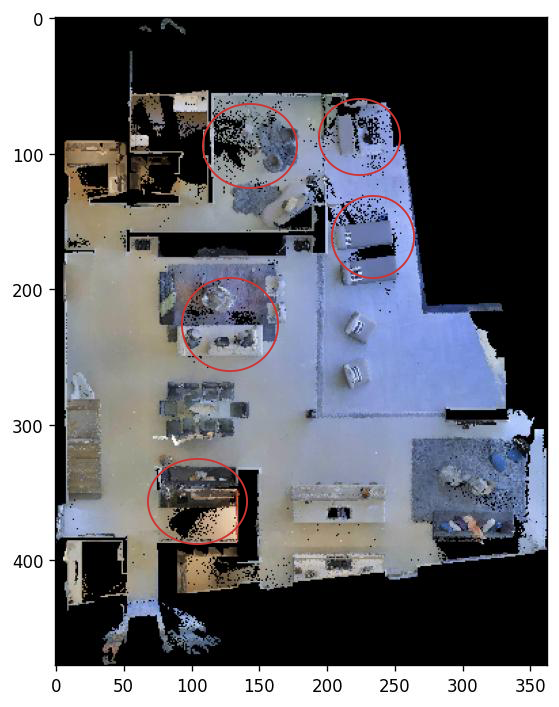}}}
\caption{3D Reconstruction Map in Bird's-Eye View}
\label{fig: 3d-restruction}
\end{figure}

Compared to other black-box data collection methods in the Habitat simulator, where predefined routes or exploration algorithms are used, this approach offers strong controllability. It allows tailored responses to the environment, enabling the collection of fewer data points while achieving superior reconstruction results. For the same scene in Matterport3D, our approach achieves comparable results to the VLMap's original authors while reducing the data volume by approximately 8\%. In certain areas, the reconstruction performance even surpasses that of the original authors. To compare the results, refer to \textcolor{red}{Fig.\ref{fig: 3d-restruction}(a)} and \textcolor{red}{Fig.\ref{fig: 3d-restruction}(b)}, for more detailed results of our 3D reconstruction bird's-eye view, please refer to \textcolor{red}{Appendix.\ref{3d-reconstruction-map}}.

% \begin{figure}[htbp]
% \centerline{\subfloat[IVLMap]{\includegraphics[width=0.24\textwidth]{images/cmu采集的3d重建图-ps强调.png}\label{fig: cmu-3d-restruction}}\ \subfloat[VLMap]{\includegraphics[width=0.24\textwidth]{images/作者的3d重建图-ps强调.png}\label{fig: vlmap-3d-restruction}}}
% \caption{3D Reconstruction Map in Bird's-Eye View}
% \label{fig: restruction}
% \end{figure}

% \begin{figure}[htbp]
%     \centering
%     \begin{minipage}{0.49\linewidth}
%         \centering
%         \includegraphics[width=0.9\linewidth]{images/cmu采集的3d重建图-ps强调.png}
%         \caption{3D Reconstruction Map in Bird's-Eye View: Ours}
%         \label{fig: cmu-3d-restruction}
%     \end{minipage}
%     %\qquad
%     \begin{minipage}{0.49\linewidth}
%         \centering
%         \includegraphics[width=0.9\linewidth]{images/作者的3d重建图-ps强调.png}
%         \caption{3D Reconstruction Map in Bird's-Eye View: VLMap}
%         \label{fig: vlmap-3d-restruction}
%     \end{minipage}
% \end{figure}

\begin{table*}[htp]\scriptsize
\centering
\newcommand{\tabincell}[2]{\begin{tabular}{@{}#1@{}}#2\end{tabular}}
\begin{tabular}{c|p{0.1cm}p{0.24cm}p{0.24cm}p{0.24cm}p{0.24cm}p{0.35cm}|p{0.1cm}p{0.24cm}p{0.24cm}p{0.24cm}p{0.24cm}p{0.35cm}|p{0.1cm}p{0.24cm}p{0.24cm}p{0.24cm}p{0.24cm}p{0.35cm}|p{0.1cm}p{0.24cm}p{0.24cm}p{0.24cm}p{0.24cm}p{0.35cm}}
\toprule
\multirow{2}{*}{Method}     &\multicolumn{6}{c|}{\tabincell{c}{5LpN3gDmAk7\\(subgoals: \textcolor{red}{28}, tasks: \textcolor{red}{7}}}  &\multicolumn{6}{c|}{\tabincell{c}{gTV8FGcVJC9\\(subgoals: \textcolor{red}{36}, tasks: \textcolor{red}{9}}}&\multicolumn{6}{c|}{\tabincell{c}{UwV83HsGsw3\\(subgoals: \textcolor{red}{40}, tasks: \textcolor{red}{10}}}&\multicolumn{6}{c}{\tabincell{c}{Vt2qJdWjCF2\\(subgoals: \textcolor{red}{40}, tasks: \textcolor{red}{10}}}\\
\cmidrule{2-25}
&SN&SR&T\_1&T\_2&T\_3&T\_4
&SN&SR&T\_1&T\_2&T\_3&T\_4
&SN&SR&T\_1&T\_2&T\_3&T\_4
&SN&SR&T\_1&T\_2&T\_3&T\_4\\
\midrule
CoW	&10&	0.36&	0.29&	0.14&	0.14&	0.00&	
13&	0.33&	0.56&	0.44&	0.17&	0.11&	
13&	0.33&	0.30&	0.10&	0.10&	0.00&	
13&	0.33&	0.30&	0.20&	0.10&	0.00\\
CLIP Map&	8&	0.29&	0.29&	0.14&	0.00&	0.00&	
11&	0.33&	0.56&	0.33&	0.11&	0.11&
11&	0.28&	0.20&	0.10&	0.00&	0.00&	
12&	0.30&	0.20&	0.10&	0.00&	0.00\\
VLMAP& 15&	0.54&	0.43&	0.29&	0.14&	0.00&
20&	0.56&	0.67&	0.56&	0.22&	0.22&
17&	0.43&	0.60&	0.30&	0.30&	0.10&
18&	0.45&	0.50&	0.40&	0.10&	0.00\\
\textbf{IVLMap(Ours)}&	\textbf{18}&	\textbf{0.64}&	\textbf{0.57}&	\textbf{0.29}&	\textbf{0.14}&	\textbf{0.00}&	
\textbf{23}&	\textbf{0.67}&	\textbf{0.78}&	\textbf{0.56}&	\textbf{0.25}&	\textbf{0.22}&	
\textbf{19}&	\textbf{0.48}&	\textbf{0.60}&	\textbf{0.40}&	\textbf{0.30}&	\textbf{0.10}&	
\textbf{20}&	\textbf{0.50}&	\textbf{0.60}&	\textbf{0.40}&	\textbf{0.20}&	\textbf{0.10}\\
\bottomrule
\end{tabular}
\caption{Outcome of Multi-Object Navigation with specified subgoals, denoted by SN for success number, SR for success rate, T\_k for achieving the kth subgoal out of the total 4 subgoals in each task, and TSR for task success rate, namely T\_4.}
\label{t2}
\end{table*}

\subsection{Multi-Object Navigation with given subgoals}
To assess the localization and navigation performance of IVLMap, we initially conducted navigation experiments with given subgoals. In the navigation experiments conducted in the four scenes of the Matterport dataset, we curated multiple navigation tasks for each scene. Each navigation task comprises four subgoals. The robot is instructed to sequentially navigate to each subgoal of each task. We use the invocation of the "stop" function by the robot as a criterion. If the robot calls the "stop" function and its distance to the target is less than a threshold (set to 1 meter in our case), it is considered successful navigation to that subgoal. Successful completion of a navigation task is achieved when the robot successfully navigates to all four subgoals in sequence. It is noteworthy that we provided instance information of objects in the given subgoals to examine the effectiveness of our constructed IVLMap.

\begin{figure}[ht]
\centerline{\subfloat[IVLMap (Ours)]{\includegraphics[width=0.20\textwidth]{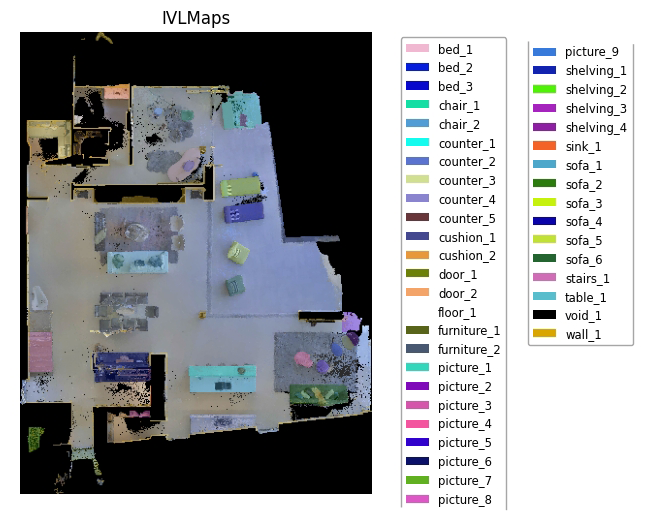}}\ \subfloat[VLMap (baseline)]{\includegraphics[width=0.20\textwidth]{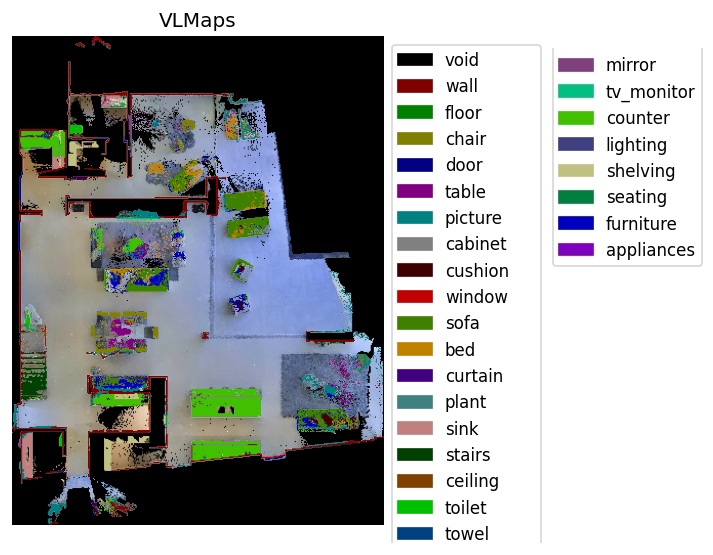}}}
\caption{Semantic segmentation results}
\label{fig: segmentation-resluts}
\end{figure}

Our observations(\textcolor{red}{Table.\ref{t2}}) indicate that our approach outperforms all other baselines. It exhibits a slight improvement in navigation performance compared to VLMap, while significantly surpassing the performance of CoW and CLIP Map. VLMap achieves zero-shot navigation by smoothly performing precise localization of landmarks. However, this baseline has limitations as it can only navigate to the nearest category to the robot agent, lacking the capability for precise instantiation navigation. As illustrated in the comparisons between \textcolor{red}{Fig.\ref{fig: segmentation-resluts}(a)} and \textcolor{red}{Fig.\ref{fig: segmentation-resluts}(b)}, our proposed IVLMap incorporates instantiation information for each landmark, enabling precise instantiation navigation tasks. Consequently, the final navigation performance is significantly enhanced. During specific localization, our approach involves initially identifying the approximate region of the landmark from the $\mathcal{U}$ and $\mathcal{V}$ matrices of IVLMap ${\mathscr M}$(\textcolor{red}{Sec.\ref{section: ivlmap-creation}}). Subsequently, further refinement is conducted using VLMap, optimizing the performance of VLMap and resulting in a significant improvement in navigation accuracy.

%please ref as follows ???\ref{fig: segmentation-resluts}\protect\subref{tt1}
% \begin{figure}[htbp]
% \centerline{\subfloat[IVLMap]{\label{fig: ivlmap}\includegraphics[width=0.24\textwidth]{images/ivlmap-优化.png}}\ \subfloat[VLMap]{\label{fig: vlmap}\includegraphics[width=0.24\textwidth]{images/vlmap-优化.png}}}
% \caption{Semantic segmentation results}
% \label{fig: segmentation}
% \end{figure}

% \begin{figure}[htbp]
%     \centering
%     \begin{minipage}{0.49\linewidth}
%         \centering
%         \includegraphics[width=0.9\linewidth]{images/ivlmap-优化.png}
%         \caption{ Semantic segmentation results: IVLMaps(Ours)}
%         \label{fig: ivlmap}
%     \end{minipage}
%     %\qquad
%     \begin{minipage}{0.49\linewidth}
%         \centering
%         \includegraphics[width=0.9\linewidth]{images/vlmap-优化.png}
%         \caption{Semantic segmentation results:VLMaps(Baseline)}
%         \label{fig: vlmap}
%     \end{minipage}
% \end{figure}

\begin{table}[!htp]\scriptsize
    \centering
    \begin{tabular}{c|cccc} 
    \toprule
         Method&  T\_1&  T\_2&  T\_3& T\_4\\
         \midrule
         CoW&  0.39&  0.22&  0.08& 0.03\\  
         CLIP Map&  0.33&  0.19&  0.06& 0.0\\   
         VLMap&  0.56&  0.39&  0.19& 0.08\\   
         \textbf{IVLMap (Ours)}&  \textbf{0.64}&  \textbf{0.44}&  \textbf{0.25}& \textbf{0.14}\\ 
         \bottomrule
    \end{tabular}
    \caption{Outcome of Zero-Shot Instance Level Object Goal Navigation from Natural Language. T\_k for achieving the kth subgoal out of the total 4 subgoals in each task. }
    \label{tab: zero-shot-navigation}
\end{table}

\subsection{Zero-Shot Instance Level Object Goal Navigation from Natural Language}
In these experiments, we assess IVLMaps' performance in comparison to alternative baselines concerning zero-shot instance-level object goal navigation initiated by natural language instructions. Our benchmark comprises 36 trajectories across four scenes, each accompanied by manually provided language instructions for evaluation purposes. In each language instruction, we provide instantiation and color information for navigation subgoals using natural language, such as "the first yellow sofa", "in between the chair and the sofa" or "east of the red table." Leveraging LLM, the robot agent extracts this information for localization and navigation. Each trajectory comprises four subgoals, and successful navigation to the proximity of a subgoal within a threshold range (set at 1m) is considered a success.

\begin{figure}[ht]
    \centering
    \includegraphics[width=0.40\textwidth]{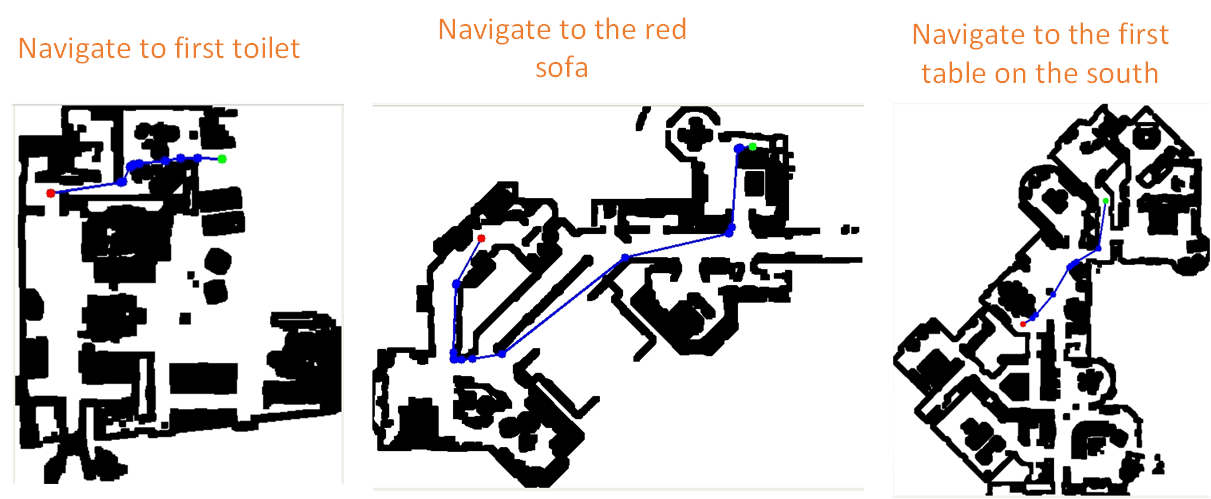}
    \caption{Zero shot navigation diagram, where green dot represents the starting point and red dot represents the endpoint. }
    \label{fig:navigation-trajectory}
    \end{figure}
Analysis of \textcolor{red}{Table.\ref{tab: zero-shot-navigation}} reveals that in zero-shot level object goal navigation, our navigation accuracy is hardly affected. This is attributed to the initial parsing of natural language instructions using LLM, enabling precise extraction of physical attributes, ensuring robust performance in navigation. Moreover, as depicted in partial trajectory schematics of navigation tasks in \textcolor{red}{Fig.\ref{fig:navigation-trajectory}}, our IVLMap achieves precise instance-level object navigation globally, a capability unmatched by other baselines.

\section{Conclusion}
% In this study, we present the Instance Level Visual Language Map (IVLMap), enhancing navigation accuracy with instance-level and attribute-level semantic language instructions. Our approach considers real-life scenarios, improving applicability. While real-world robot applications show promising initial results, mapping performance in such environments needs enhancement for better navigation. The static nature of our maps in dynamic real-world environments requires improvement, suggesting the integration of real-time navigation using laser scanners. Additionally, we aim to advance towards 3D semantic maps, enabling dynamic perception of object height for accurate spatial navigation. Continuous research efforts will address these challenges. 

In this study, we introduce the Instance Level Visual Language Map (IVLMap), elevating navigation precision through instance-level and attribute-level semantic language instructions. Our approach is designed to enhance applicability in real-life scenarios, showing promising results in initial real-world robot applications. However, the mapping performance in dynamic environments requires improvement, prompting the exploration of real-time navigation using laser scanners. Our future goals include advancing towards 3D semantic maps to enable dynamic perception of object height, contributing to more accurate spatial navigation. Ongoing research efforts will focus on addressing these challenges.

\section*{Acknowledgment}
The work is supported by the National Natural Science Foundation of China (no. 61601112). It is also supported by the Fundamental Research Funds for the Central Universities and DHU Distinguished Young Professor Program.

\bibliographystyle{IEEEtranDOIISBNISSN}
%\bibliography{paper.bib}

\begin{thebibliography}{10}
\providecommand{\url}[1]{#1}
\csname url@samestyle\endcsname
\providecommand{\newblock}{\relax}
\providecommand{\bibinfo}[2]{#2}
\providecommand{\BIBentrySTDinterwordspacing}{\spaceskip=0pt\relax}
\providecommand{\BIBentryALTinterwordstretchfactor}{4}
\providecommand{\BIBentryALTinterwordspacing}{\spaceskip=\fontdimen2\font plus
\BIBentryALTinterwordstretchfactor\fontdimen3\font minus
  \fontdimen4\font\relax}
\providecommand{\BIBforeignlanguage}[2]{{%
\expandafter\ifx\csname l@#1\endcsname\relax
\typeout{** WARNING: IEEEtran.bst: No hyphenation pattern has been}%
\typeout{** loaded for the language `#1'. Using the pattern for}%
\typeout{** the default language instead.}%
\else
\language=\csname l@#1\endcsname
\fi
#2}}
\providecommand{\BIBdecl}{\relax}
\BIBdecl

\bibitem{fu2024mobile}
Z.~Fu, T.~Z. Zhao, and C.~Finn, ``Mobile aloha: Learning bimanual mobile
  manipulation with low-cost whole-body teleoperation,'' in \emph{arXiv}, 2024.

\bibitem{anderson2018vision}
P.~Anderson, Q.~Wu, D.~Teney, J.~Bruce, M.~Johnson, N.~S{\"u}nderhauf, I.~Reid,
  S.~Gould, and A.~Van Den~Hengel, ``Vision-and-language navigation:
  Interpreting visually-grounded navigation instructions in real
  environments,'' in \emph{Proceedings of the IEEE conference on computer
  vision and pattern recognition}, 2018, pp. 3674--3683.

\bibitem{gu2022vision}
J.~Gu, E.~Stefani, Q.~Wu, J.~Thomason, and X.~E. Wang, ``Vision-and-language
  navigation: A survey of tasks, methods, and future directions,'' \emph{arXiv
  preprint arXiv:2203.12667}, 2022.

\bibitem{shah2023lm}
D.~Shah, B.~Osi{\'n}ski, S.~Levine \emph{et~al.}, ``Lm-nav: Robotic navigation
  with large pre-trained models of language, vision, and action,'' in
  \emph{Conference on Robot Learning}.\hskip 1em plus 0.5em minus 0.4em\relax
  PMLR, 2023, pp. 492--504.

\bibitem{huang2023visual}
C.~Huang, O.~Mees, A.~Zeng, and W.~Burgard, ``Visual language maps for robot
  navigation,'' in \emph{2023 IEEE International Conference on Robotics and
  Automation (ICRA)}.\hskip 1em plus 0.5em minus 0.4em\relax IEEE, 2023, pp.
  10\,608--10\,615.

\bibitem{qi2020reverie}
Y.~Qi, Q.~Wu, P.~Anderson, X.~Wang, W.~Y. Wang, C.~Shen, and A.~v.~d. Hengel,
  ``Reverie: Remote embodied visual referring expression in real indoor
  environments,'' in \emph{Proceedings of the IEEE/CVF Conference on Computer
  Vision and Pattern Recognition}, 2020, pp. 9982--9991.

\bibitem{zhu2021soon}
F.~Zhu, X.~Liang, Y.~Zhu, Q.~Yu, X.~Chang, and X.~Liang, ``Soon: Scenario
  oriented object navigation with graph-based exploration,'' in
  \emph{Proceedings of the IEEE/CVF Conference on Computer Vision and Pattern
  Recognition}, 2021, pp. 12\,689--12\,699.

\bibitem{wang2023scaling}
Z.~Wang, J.~Li, Y.~Hong, Y.~Wang, Q.~Wu, M.~Bansal, S.~Gould, H.~Tan, and
  Y.~Qiao, ``Scaling data generation in vision-and-language navigation,'' in
  \emph{Proceedings of the IEEE/CVF International Conference on Computer
  Vision}, 2023, pp. 12\,009--12\,020.

\bibitem{he2021landmark}
K.~He, Y.~Huang, Q.~Wu, J.~Yang, D.~An, S.~Sima, and L.~Wang, ``Landmark-rxr:
  Solving vision-and-language navigation with fine-grained alignment
  supervision,'' \emph{Advances in Neural Information Processing Systems},
  vol.~34, pp. 652--663, 2021.

\bibitem{kirillov2023segment}
A.~Kirillov, E.~Mintun, N.~Ravi, H.~Mao, C.~Rolland, L.~Gustafson, T.~Xiao,
  S.~Whitehead, A.~C. Berg, W.-Y. Lo \emph{et~al.}, ``Segment anything,''
  \emph{arXiv preprint arXiv:2304.02643}, 2023.

\bibitem{fukushima1980neocognitron}
K.~Fukushima, ``Neocognitron: A self-organizing neural network model for a
  mechanism of pattern recognition unaffected by shift in position,''
  \emph{Biological cybernetics}, vol.~36, no.~4, pp. 193--202, 1980.

\bibitem{estefo2019robot}
P.~Estefo, J.~Simmonds, R.~Robbes, and J.~Fabry, ``The robot operating system:
  Package reuse and community dynamics,'' \emph{Journal of Systems and
  Software}, vol. 151, pp. 226--242, 2019.

\bibitem{salas2013slam++}
R.~F. Salas-Moreno, R.~A. Newcombe, H.~Strasdat, P.~H. Kelly, and A.~J.
  Davison, ``Slam++: Simultaneous localisation and mapping at the level of
  objects,'' in \emph{Proceedings of the IEEE conference on computer vision and
  pattern recognition}, 2013, pp. 1352--1359.

\bibitem{mccormac2018fusion++}
J.~McCormac, R.~Clark, M.~Bloesch, A.~Davison, and S.~Leutenegger, ``Fusion++:
  Volumetric object-level slam,'' in \emph{2018 international conference on 3D
  vision (3DV)}.\hskip 1em plus 0.5em minus 0.4em\relax IEEE, 2018, pp. 32--41.

\bibitem{chen2023open}
B.~Chen, F.~Xia, B.~Ichter, K.~Rao, K.~Gopalakrishnan, M.~S. Ryoo, A.~Stone,
  and D.~Kappler, ``Open-vocabulary queryable scene representations for real
  world planning,'' in \emph{2023 IEEE International Conference on Robotics and
  Automation (ICRA)}.\hskip 1em plus 0.5em minus 0.4em\relax IEEE, 2023, pp.
  11\,509--11\,522.

\bibitem{zhang2019pose2seg}
S.-H. Zhang, R.~Li, X.~Dong, P.~Rosin, Z.~Cai, X.~Han, D.~Yang, H.~Huang, and
  S.-M. Hu, ``Pose2seg: Detection free human instance segmentation,'' in
  \emph{Proceedings of the IEEE/CVF conference on computer vision and pattern
  recognition}, 2019, pp. 889--898.

\bibitem{lee2020centermask}
Y.~Lee and J.~Park, ``Centermask: Real-time anchor-free instance
  segmentation,'' in \emph{Proceedings of the IEEE/CVF conference on computer
  vision and pattern recognition}, 2020, pp. 13\,906--13\,915.

\bibitem{long2023discuss}
Y.~Long, X.~Li, W.~Cai, and H.~Dong, ``Discuss before moving: Visual language
  navigation via multi-expert discussions,'' \emph{arXiv preprint
  arXiv:2309.11382}, 2023.

\bibitem{jia2022vital}
Z.~Jia, K.~Yu, J.~Ru, S.~Yang, and S.~Coleman, ``Vital information matching in
  vision-and-language navigation,'' \emph{Frontiers in Neurorobotics}, vol.~16,
  p. 1035921, 2022.

\bibitem{vasudevan2021talk2nav}
A.~B. Vasudevan, D.~Dai, and L.~Van~Gool, ``Talk2nav: Long-range
  vision-and-language navigation with dual attention and spatial memory,''
  \emph{International Journal of Computer Vision}, vol. 129, pp. 246--266,
  2021.

\bibitem{chen2022weakly}
P.~Chen, D.~Ji, K.~Lin, R.~Zeng, T.~Li, M.~Tan, and C.~Gan, ``Weakly-supervised
  multi-granularity map learning for vision-and-language navigation,''
  \emph{Advances in Neural Information Processing Systems}, vol.~35, pp.
  38\,149--38\,161, 2022.

\bibitem{krantz2023iterative}
J.~Krantz, S.~Banerjee, W.~Zhu, J.~Corso, P.~Anderson, S.~Lee, and J.~Thomason,
  ``Iterative vision-and-language navigation,'' in \emph{Proceedings of the
  IEEE/CVF Conference on Computer Vision and Pattern Recognition}, 2023, pp.
  14\,921--14\,930.

\bibitem{zhou2023navgpt}
G.~Zhou, Y.~Hong, and Q.~Wu, ``Navgpt: Explicit reasoning in
  vision-and-language navigation with large language models,'' \emph{arXiv
  preprint arXiv:2305.16986}, 2023.

\bibitem{schumann2023velma}
R.~Schumann, W.~Zhu, W.~Feng, T.-J. Fu, S.~Riezler, and W.~Y. Wang, ``Velma:
  Verbalization embodiment of llm agents for vision and language navigation in
  street view,'' \emph{arXiv preprint arXiv:2307.06082}, 2023.

\bibitem{vemprala2023chatgpt}
S.~Vemprala, R.~Bonatti, A.~Bucker, and A.~Kapoor, ``Chatgpt for robotics:
  Design principles and model abilities. 2023,'' \emph{Published by Microsoft},
  2023.

\bibitem{driess2023palm}
D.~Driess, F.~Xia, M.~S. Sajjadi, C.~Lynch, A.~Chowdhery, B.~Ichter, A.~Wahid,
  J.~Tompson, Q.~Vuong, T.~Yu \emph{et~al.}, ``Palm-e: An embodied multimodal
  language model,'' \emph{arXiv preprint arXiv:2303.03378}, 2023.

\bibitem{Li_Weinberger_Belongie_Koltun_Ranftl}
B.~Li, K.~Weinberger, S.~Belongie, V.~Koltun, and R.~Ranftl,
  ``\BIBforeignlanguage{en-US}{Language-driven semantic segmentation},'' 2023.

\bibitem{ray2023chatgpt}
P.~P. Ray, ``Chatgpt: A comprehensive review on background, applications, key
  challenges, bias, ethics, limitations and future scope,'' \emph{Internet of
  Things and Cyber-Physical Systems}, 2023.

\bibitem{touvron2023llama}
H.~Touvron, L.~Martin, K.~Stone, P.~Albert, A.~Almahairi, Y.~Babaei,
  N.~Bashlykov, S.~Batra, P.~Bhargava, S.~Bhosale \emph{et~al.}, ``Llama 2:
  Open foundation and fine-tuned chat models,'' \emph{arXiv preprint
  arXiv:2307.09288}, 2023.

\bibitem{gadre2022clip}
S.~Y. Gadre, M.~Wortsman, G.~Ilharco, L.~Schmidt, and S.~Song, ``Clip on
  wheels: Zero-shot object navigation as object localization and exploration,''
  \emph{arXiv preprint arXiv:2203.10421}, vol.~3, no.~4, p.~7, 2022.

\bibitem{ahn2022can}
M.~Ahn, A.~Brohan, N.~Brown, Y.~Chebotar, O.~Cortes, B.~David, C.~Finn, C.~Fu,
  K.~Gopalakrishnan, K.~Hausman \emph{et~al.}, ``Do as i can, not as i say:
  Grounding language in robotic affordances,'' \emph{arXiv preprint
  arXiv:2204.01691}, 2022.

\bibitem{zeng2022socratic}
A.~Zeng, M.~Attarian, B.~Ichter, K.~Choromanski, A.~Wong, S.~Welker,
  F.~Tombari, A.~Purohit, M.~Ryoo, V.~Sindhwani \emph{et~al.}, ``Socratic
  models: Composing zero-shot multimodal reasoning with language,'' \emph{arXiv
  preprint arXiv:2204.00598}, 2022.

\bibitem{liang2023code}
J.~Liang, W.~Huang, F.~Xia, P.~Xu, K.~Hausman, B.~Ichter, P.~Florence, and
  A.~Zeng, ``Code as policies: Language model programs for embodied control,''
  in \emph{2023 IEEE International Conference on Robotics and Automation
  (ICRA)}.\hskip 1em plus 0.5em minus 0.4em\relax IEEE, 2023, pp. 9493--9500.

\bibitem{frantar2022gptq}
E.~Frantar, S.~Ashkboos, T.~Hoefler, and D.~Alistarh, ``Gptq: Accurate
  post-training quantization for generative pre-trained transformers,''
  \emph{arXiv preprint arXiv:2210.17323}, 2022.

\bibitem{savva2019habitat}
M.~Savva, A.~Kadian, O.~Maksymets, Y.~Zhao, E.~Wijmans, B.~Jain, J.~Straub,
  J.~Liu, V.~Koltun, J.~Malik \emph{et~al.}, ``Habitat: A platform for embodied
  ai research,'' in \emph{Proceedings of the IEEE/CVF international conference
  on computer vision}, 2019, pp. 9339--9347.

\bibitem{selvaraju2017grad}
R.~R. Selvaraju, M.~Cogswell, A.~Das, R.~Vedantam, D.~Parikh, and D.~Batra,
  ``Grad-cam: Visual explanations from deep networks via gradient-based
  localization,'' in \emph{Proceedings of the IEEE international conference on
  computer vision}, 2017, pp. 618--626.

\bibitem{schumann2022analyzing}
R.~Schumann and S.~Riezler, ``Analyzing generalization of vision and language
  navigation to unseen outdoor areas,'' \emph{arXiv preprint arXiv:2203.13838},
  2022.

\bibitem{dai2017scannet}
A.~Dai, A.~X. Chang, M.~Savva, M.~Halber, T.~Funkhouser, and M.~Nie{\ss}ner,
  ``Scannet: Richly-annotated 3d reconstructions of indoor scenes,'' in
  \emph{Proceedings of the IEEE conference on computer vision and pattern
  recognition}, 2017, pp. 5828--5839.

\bibitem{geiger2013vision}
A.~Geiger, P.~Lenz, C.~Stiller, and R.~Urtasun, ``Vision meets robotics: The
  kitti dataset,'' \emph{The International Journal of Robotics Research},
  vol.~32, no.~11, pp. 1231--1237, 2013.

\bibitem{cao2022autonomous}
C.~Cao, H.~Zhu, F.~Yang, Y.~Xia, H.~Choset, J.~Oh, and J.~Zhang, ``Autonomous
  exploration development environment and the planning algorithms,'' in
  \emph{2022 International Conference on Robotics and Automation (ICRA)}.\hskip
  1em plus 0.5em minus 0.4em\relax IEEE, 2022, pp. 8921--8928.

\end{thebibliography}

%\clearpage
\section{appendix}
\appendix

\subsection{Full List of Navigation High-Level Functions }
\label{appendix: high-level-functions}
On the basis of VLMap, we have developed a new high-level function library based on the distinctive features of IVLMap. Our full list of navigation high-level functions are listed in Table. \ref{tab:functions}

% \usepackage{booktabs}
\begin{table}\scriptsize
\centering
\begin{tabular}{p{4.4cm}|p{3.6cm}}
\toprule
\textbf{high-level functions}                     & \textbf{function description}                                            \\ \midrule \rowcolor{gray!20}get\_nearest\_obj\_pos(object\_name) & get the map position of the nearest front object. \\
 get\_obj\_attributes(object\_name, instance\_idx, objext\_color)&get the object attributes in Tuple( object\_name,label\_id,color)from     IVLMap\\
\rowcolor{gray!20} get\_specifed\_obj\_pos(object\_with\_attributes) & get the map position of the object with specifed attributes.             \\ 
get\_nearest\_obj\_contour(object\_name)          & get the contour turning pointsof the nearest front obiect on the map.    \\ 
\rowcolor{gray!20} move\_to(pos)                                     & move to a position on the map.                                           \\ 
move\_to\_left(objedt\_with\_attributes)          & move to the left side of the object with specified attributes.           \\ 
\rowcolor{gray!20} move\_to\_right(object\_with\_attributes)         & move to the right side of the object with specified attributes.          \\ 
with object on left(object\_with\_attributes) & turn until the specific object is on the robot's left side.              \\ 
\rowcolor{gray!20} with object on right(object\_with\_attributes)   & turn until the specific obiect is on the robot's right side.             \\ 
move\_in\_between(obiect\_a, object\_b)           & move in between two objects.                                             \\ 
\rowcolor{gray!20} turn(angle)                                       & turn right a certain angle. If the angle value is negative, turn left.   \\
face(object\_with\_attributes)                    & turn until the object is with specific attributes in front of the robot. \\ 
\rowcolor{gray!20} turn\_absolute(angle)                             & turn to absolute angle. 0 isnorth, 90 is east, -90 is west,180 is south. \\ 
move\_north(object\_with\_attributes)             & move to the north side of the object with specific attributes.           \\ 
\rowcolor{gray!20} move\_south(object\_with\_attributes)             & move to the south side of the object with specific attributes.           \\ 
move\_east(object\_with\_attributes)              & move to the east side of the object with specific attributes.            \\ 
\rowcolor{gray!20} move\_west(object\_with\_attributes)              & move to the west side of the object with specific attributes.            \\ 
move\_to\_obiect(object\_with\_attributes)        & move to the object with specific attributes.                             \\ 
\rowcolor{gray!20} move\_forward(dist)                               & move forward dist' meters.                                               \\ \bottomrule
\end{tabular}
\caption{List of High-level functions Used}
\label{tab:functions} % 添加标签
\end{table}

\subsection{Full Prompts for LLM Generating Python Code}
\label{appendix: full-prompts}
\textbf{system\_prompt}

\begin{tcolorbox}[colback=green!3!white,colframe=black!30!white]
\hspace*{-1.5em} \begin{tabular}{m{22.8em}} 
You are an assistant helping me with the mobile robot agent.
When I ask you to do something, you are supposed to give me Python code that is needed to achieve that task with explanations of what that code does.
You are only allowed to use the functions I have defined for you.
You are not to use any other hypothetical functions that you think might exist.
You can use simple Python functions from libraries such as math and numpy.
\end{tabular}
\label{system-prompt}
\end{tcolorbox} 

\textbf{attributes\_prompt}

\begin{tcolorbox}[colback=green!3!white,colframe=black!30!white,breakable]
\hspace*{-1.5em} \begin{tabular}{m{23.1em}} 
You should extract the object attributes from uer's command, here are some examples:
\end{tabular}
\hspace*{-1.5em} \begin{tabular}{m{1em}m{20.6em}} 
\rowcolor{blue!10}
I\ :& navigate to the third yellow table.  
\end{tabular}\\
\hspace*{-1.5em} \begin{tabular}{m{1em}m{20.6em}} 
A:& [(table,third,yellow)]
\end{tabular}
\hspace*{-1.5em} \begin{tabular}{m{1em}m{20.6em}} 
\rowcolor{blue!10}
I\ :& go to the nearest yellow chair and then go to the black sofa.  
\end{tabular}
\hspace*{-1.5em} \begin{tabular}{m{1em}m{20.6em}}  
A: &[(chair,None,yellow),(sofa,None,black)]
\end{tabular}
\hspace*{-1.5em} \begin{tabular}{m{1em}m{20.6em}}   
\rowcolor{blue!10}
I\ :& go to the kitchen and then go to the toilet. 
\end{tabular}
\hspace*{-1.5em} \begin{tabular}{m{1em}m{20.6em}}  
A:& [(kitchen,None,None), (toilet,None,None)]
\end{tabular}
\hspace*{-1.5em} \begin{tabular}{m{1em}m{20.6em}}  
\rowcolor{blue!10}
I\ :& Move to the west of the black chair, with the first red sofa on your right, move to the 2nd table, then turn right 90 degree, then find a table.  
\end{tabular}
\hspace*{-1.5em} \begin{tabular}{m{1em}m{20.6em}}  
A:& [(chair, None, black), (sofa, first, red), (table, sencond, None), (table, None, None)]
\end{tabular}
\label{attributes-prompt}
\end{tcolorbox} 

\textbf{function\_prompt}

\begin{tcolorbox}[colback=green!3!white,colframe=black!30!white]
\hspace*{-1.5em} \begin{tabular}{m{22.5em}} 
Here are some functions you can use to command the mobile robot agent.\\
\textcolor{brown}{\# get the map position of the nearest front object.}
\end{tabular}
\hspace*{-1.5em} \begin{tabular}{m{22.8em}} 
\rowcolor{blue!10}
agent.get\_nearest\_obj\_pos(object\_name)
\end{tabular}
\hspace*{-1.5em} \begin{tabular}{m{22.8em}} 
\textcolor{brown}{\# get the object attributes in Tuple(ob-ject name,label id,color)from IVLMap}
\end{tabular}
\hspace*{-1.5em} \begin{tabular}{m{22.8em}} 
\rowcolor{blue!10}
agent.get\_obj\_attributes(object\_name,instance\_idx, objext\_color)\\
\end{tabular}
\hspace*{-1.5em} \begin{tabular}{m{22.8em}} 
...\\
\textcolor{brown}{\# move forward dist' meters.If the angle value is negative, move backward.}
\end{tabular}
\hspace*{-1.5em} \begin{tabular}{m{22.8em}} 
\rowcolor{blue!10}
agent.move\_forward(dist)
\end{tabular}
\label{function-prompt}
\end{tcolorbox} 
The omitted function prompt can be referred to in \textcolor{red}{Table \ref{tab:functions}}.

\textbf{example\_prompt}

\begin{tcolorbox}[colback=green!3!white,colframe=black!30!white]
\hspace*{-1.5em} \begin{tabular}{m{22.6em}} 
Here are some examples.
\end{tabular}
\hspace*{-1.5em} \begin{tabular}{m{1em}m{20.6em}} 
\rowcolor{blue!10}
Me:& navigate to the third yellow table
\end{tabular}
\hspace*{-1.5em} \begin{tabular}{m{1em}m{20.6em}}  
You:& obj\_attr = agent.get\_obj\_attributes('table',3,'yellow')\\ & agent.face(obj\_attr)
\end{tabular}
\hspace*{-1.5em} \begin{tabular}{m{1em}m{20.6em}}      
\rowcolor{blue!10}
Me:& Move to the red sofa then stop at the right side of it
\end{tabular}
\hspace*{-1.5em} \begin{tabular}{m{1em}m{20.6em}}       
You:& obj\_attr = agent.get\_obj\_attributes('sofa',0,'red')\\ & agent.move\_to\_right(obj\_attr)
\end{tabular}
\hspace*{-1.5em} \begin{tabular}{m{1em}m{20.6em}}        
\rowcolor{blue!10}
Me:& Go straight for 1 meter and then walk clockwise around the red sofa for 2 turns.
\end{tabular}
\hspace*{-1.5em} \begin{tabular}{m{1em}m{20.6em}}     
You:& agent.move\_forward(1)\\
&obj\_attr = agent.get\_obj\_attributes( 'sofa', 0, 'red' )\\ 
&obj\_contour = agent.get\_specifed\_obj\_pos(obj\_attr)\\
& agent.face(obj\_attr), agent.turn(-90)\\
 &for i in range(8):\\
&\quad agent.move\_forward(obj\_contour([i\%4])\\
&\quad agent.turn(90)
\end{tabular}
\label{example-prompt}
\end{tcolorbox} 

\textbf{gen\_code\_prompt}

\begin{tcolorbox}
[colback=green!3!white,colframe=black!30!white]
\hspace*{-1.5em} \begin{tabular}{m{22.6em}} 
here are some examples for you to generate code in python according to Human's command. You should return your answer in Markdown format, especially when wrapping Python code with '''python    ''':
\end{tabular}
\hspace*{-1.5em} \begin{tabular}{m{3em}m{18.6em}}
\rowcolor{blue!10}
Human:& navigate to the third yellow table
\end{tabular}
\hspace*{-1.5em} \begin{tabular}{m{3em}m{18.6em}} 
Assistant:&'''python\\
&obj\_attr = agent.get\_obj\_attributes('table',3,'yel\\
&low')\\ 
&agent.face(obj\_attr)\\
&'''
\end{tabular}
\label{gen-code-prompt}
\end{tcolorbox}

\subsection{Experiment on Real Robot}
\label{appendix: experiment-on-real-robot}
\begin{figure}[ht]
    \centering
    \includegraphics[width=0.40\textwidth]{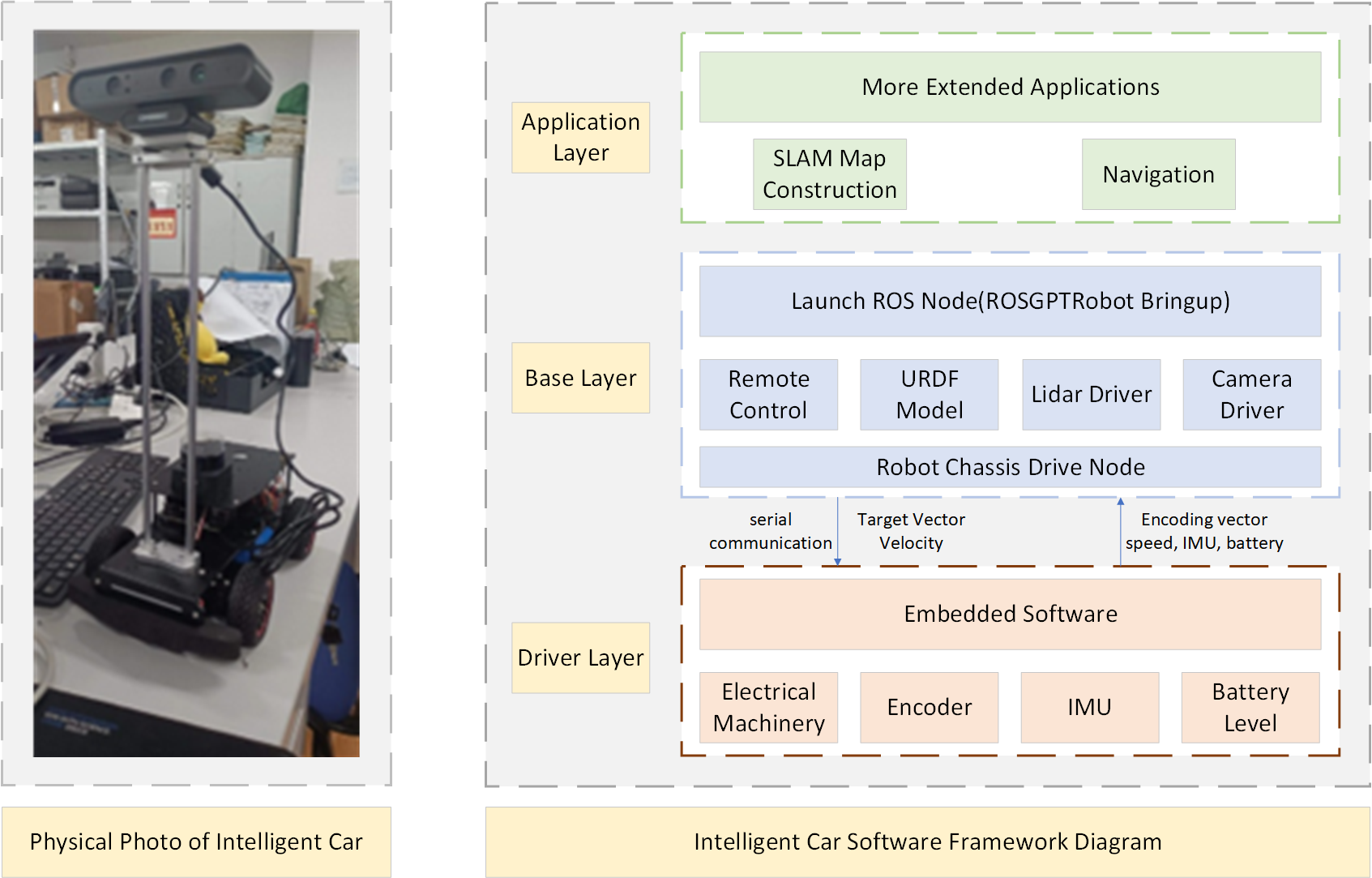}
    \caption{We built an intelligent robotic car with ROS, leveraging Jetson Nano for deep learning. It includes a high-precision Orbbec Astra Pro Plus monocular depth camera and Slamtec E300 LiDAR for RGB-D and LiDAR data capture.}
    \label{fig:dataset-base-on-real-environment}
    \end{figure}

\textbf{ROS-based Smart Car Real-World Data Collection Scheme: } In current research on visual language navigation, the majority of work is implemented using simulators such as Habitat and AI2THOR, achieving notable results in virtual environments. To validate the effectiveness of the algorithm in real-world scenarios, we conducted corresponding experiments in actual environments. Our real-world data collection platform is illustrated in \textcolor{red}{Fig.\ref{fig:dataset-base-on-real-environment}}. Before initiating the data collection process, we performed camera calibration to establish the transformation relationship between the robot base coordinate system and the camera coordinate system. During the data collection process, it is crucial to ensure that the robot and the host are on the same local network. The robot's movement is controlled using the laptop keyboard through the ROS communication mechanism. RGB and depth information is captured through the Astro Pro Plus camera, while the pose information is obtained from IMU and the robot's velocity encoder. These sensor data are published as ROS topics. Subsequently, ROS \textcolor{blue}{message\_filter\footnote{ROS message\_filter is a library in the Robot Operating System (ROS) that provides tools for synchronizing and filtering messages from multiple topics in a flexible and efficient manner. }} is utilized to synchronize the three different types of sensor data, ensuring they are roughly at the same timestamp. 

\begin{figure}[h]
    \centering
    \includegraphics[width=0.40\textwidth]{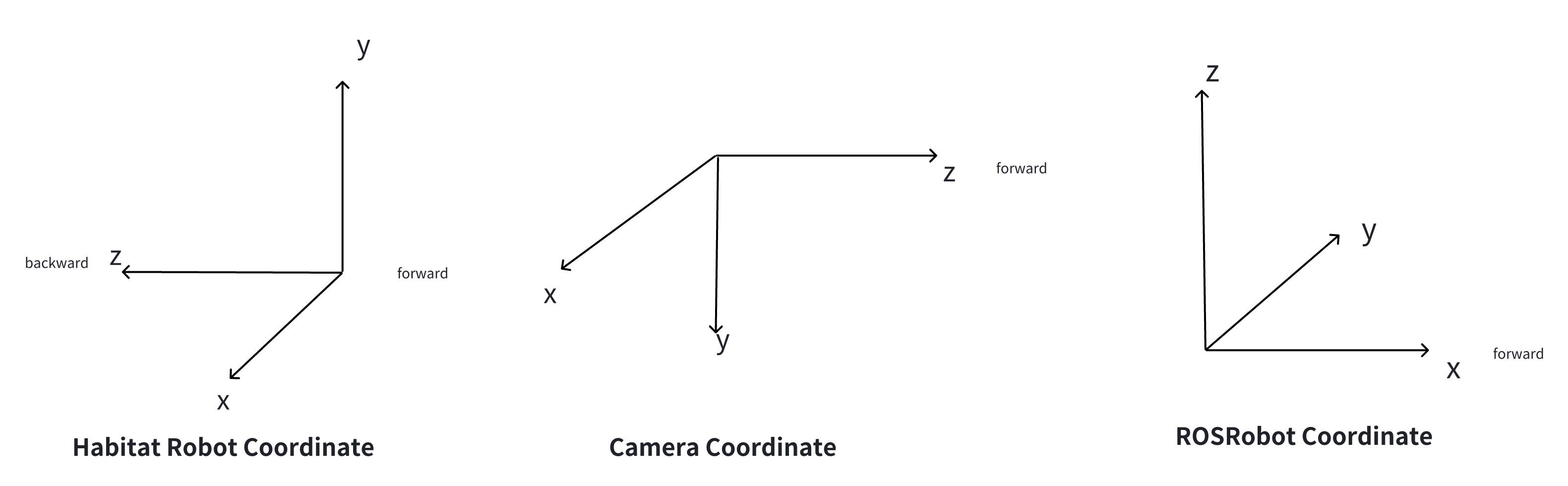}
    \caption{Habitat coordinate system, camera coordinate system, and ROSRobot coordinate system }
    \label{fig:coordinate}
    \end{figure}

In the real-world environment, where only odometer information reflecting the pose changes of the robot car is available, a coordinate transformation between the ROSRobot base coordinate system and the Camera coordinate system is necessary(\textcolor{red}{Fig.\ref{fig:coordinate}}). The rotational relationship between the robot coordinate system and the camera coordinate system can be expressed using the rotation matrix shown in \textcolor{red}{Equation.\ref{eq: rotation}}. 
\begin{equation}
    r o t=\left[\begin{array}{ccc}
0 & 0 & 1 \\
-1 & 0 & 0 \\
0 & -1 & 0
\end{array}\right]
\label{eq: rotation}
\end{equation}

\begin{figure}[h]
    \centering
    \includegraphics[width=0.40\textwidth]{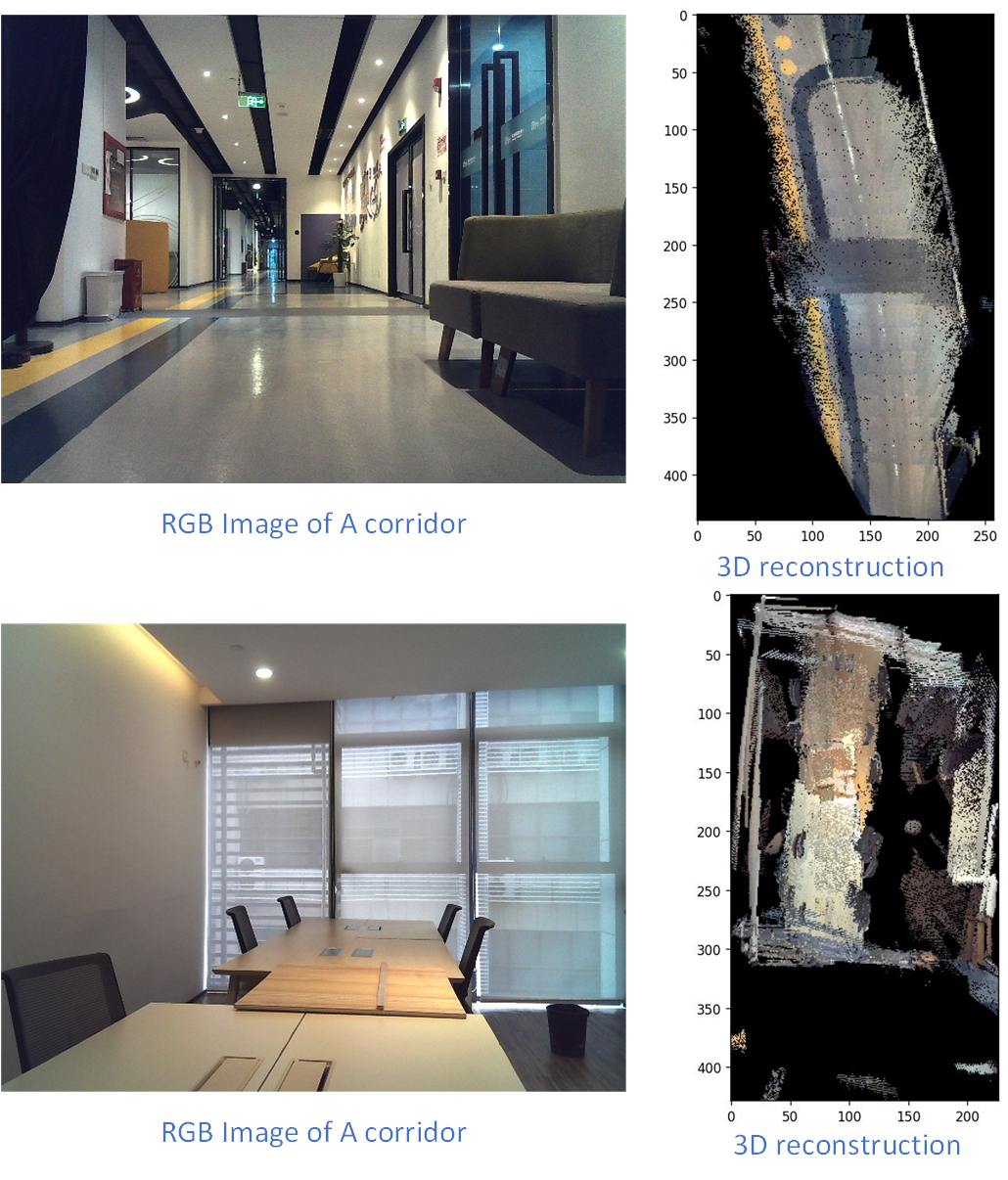}
    \caption{3D Reconstruction Results in Real Environment,  examples of RGB images captured by the camera on the left, the corresponding 3D reconstruction results are displayed on the right.}
    \label{fig:real-3d-reconstruction-with-rgb}
    \end{figure}

\subsection{3D Reconstruction  in Bird's-Eye View}
\label{3d-reconstruction-map}
\begin{figure*}[htbp]%调节图片位置，h：浮动；t：顶部；b:底部；p：当前位置
	\centering
	\includegraphics[width=0.9\linewidth]{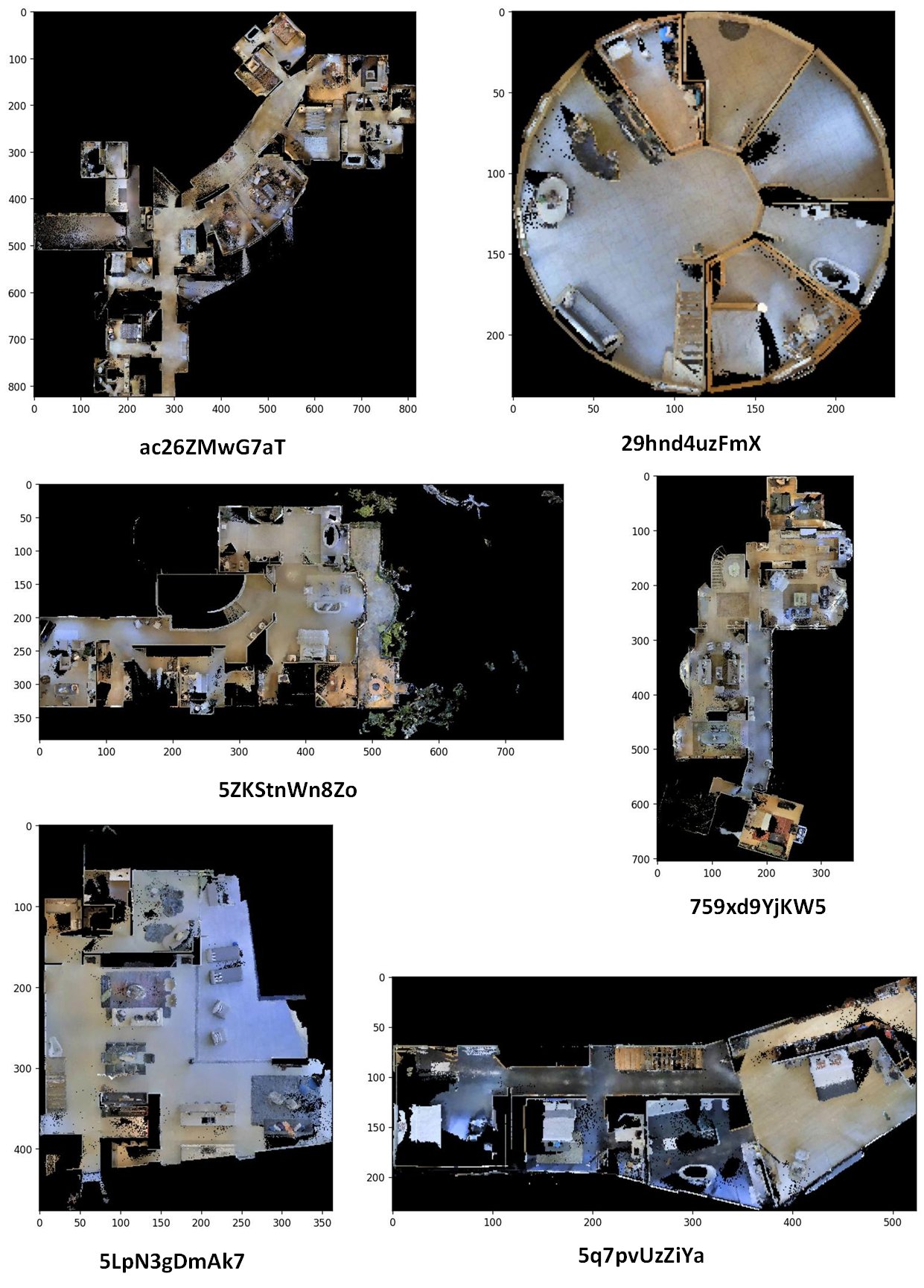}% 中括号中的为调节图片大小
	\caption{3D Reconstruction Bird's-Eye View of Different Scenes}
	\label{cmu-3d-reconstructions}%文中引用该图片代号
\end{figure*}

3D Reconstruction Bird's-Eye View of Different Scenes in the Dataset Collected with cmu-exploration can be refered at \textcolor{red}{Fig. \ref{cmu-3d-reconstructions}}.

\subsection{IVLMap segmentation results}
\label{appendix: ivlmap-vs-vlmap}
\begin{figure*}[htbp]%调节图片位置，h：浮动；t：顶部；b:底部；p：当前位置
	\centering
	\includegraphics[width=0.8\linewidth]{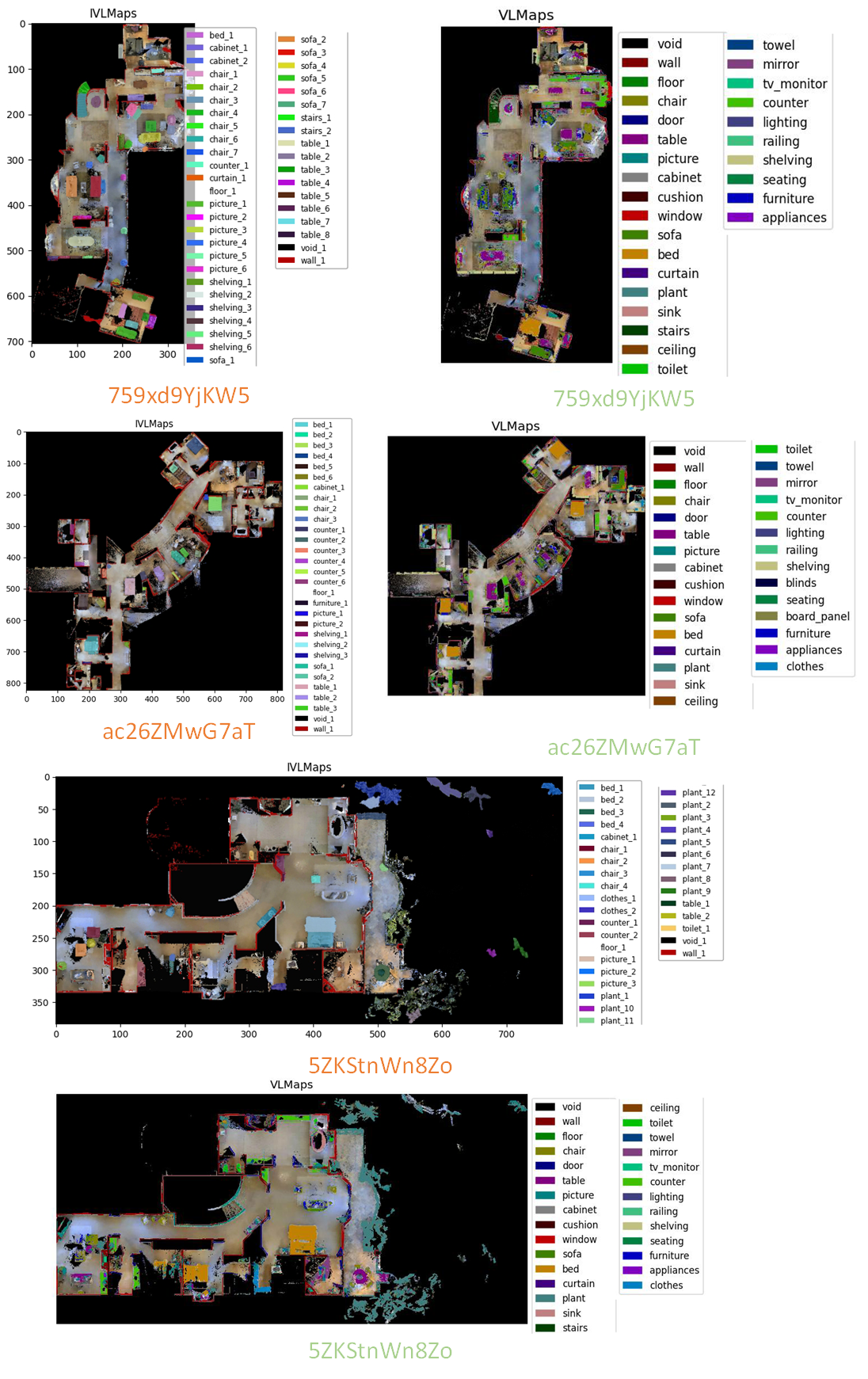}% 中括号中的为调节图片大小
	\caption{Semantic Map Comparison between IVLMap (Ours, Orange) and VLMap (GT, Green)}
	\label{ivlmap-vs-vlmap}%文中引用该图片代号
\end{figure*}

To compare the semantic maps between IVLMap (ours, depicted in orange) and VLMap (ground truth, represented in green), refer to the visual representation of the maps for a comprehensive understanding of the differences and similarities  \textcolor{red}{Fig.\ref{ivlmap-vs-vlmap}}.

\end{document}